\documentclass[preprint,12pt]{elsarticle}

\usepackage{amssymb}
\usepackage{amsmath}
\usepackage{enumitem}
\usepackage{subcaption}%
\usepackage{adjustbox} %c
\usepackage{makecell}
\usepackage{url}
\usepackage{float}
\usepackage{multirow}%

\usepackage{booktabs}
\newcommand{\tabitem}{~~\llap{\textbullet}~~}
\usepackage{tabularx}

\usepackage{lmodern}
\usepackage{color}
\usepackage{soul}

\journal{arXiv}

\begin{document}

\begin{frontmatter}



\title{REVEX: A Unified Framework for Removal-Based Explainable Artificial Intelligence in Video}


\author[1,2,3]{F. Xavier Gaya-Morey\corref{cor1}}
\ead{francesc-xavier.gaya@uib.es}
\author[1,2,3]{Jose M. Buades-Rubio}
\ead{josemaria.buades@uib.es}
\author[4]{I. Scott MacKenzie}
\ead{mack@yorku.ca}
\author[1,2,3]{Cristina Manresa-Yee}
\ead{cristina.manresa@uib.es}

\cortext[cor1]{Corresponding author.}

\affiliation[1]{%
    organization={Computer Graphics and Vision and AI Group (UGIVIA), Universitat de les Illes Balears},
    addressline={Carretera de Valldemossa, km 7.5}, 
    city={Palma, Illes Balears},
    postcode={07122}, 
    country={Spain}}
\affiliation[2]{%
    organization={Research Institute of Health Sciences (IUNICS), Universitat de les Illes Balears},
    addressline={Carretera de Valldemossa, km 7.5}, 
    city={Palma, Illes Balears},
    postcode={07122}, 
    country={Spain}}
\affiliation[3]{%
    organization={Department of Mathematics and Computer Science, Universitat de les Illes Balears},
    addressline={Carretera de Valldemossa, km 7.5}, 
    city={Palma, Illes Balears},
    postcode={07122}, 
    country={Spain}}
\affiliation[4]{%
    organization={Department of Electrical Engineering and Computer Science, York University},
    addressline={4700 Keele St}, 
    city={Toronto},
    postcode={M3J 1P3}, 
    country={Canada}}

% \affiliation{organization={},%Department and Organization

%% Abstract
\begin{abstract}
We developed REVEX, a \underline{re}moval-based \underline{v}ideo \underline{ex}planations framework. This work extends fine-grained explanation frameworks for computer vision data and adapts six existing techniques to video by adding temporal information and local explanations. The adapted methods were evaluated across networks, datasets, image classes, and evaluation metrics. By decomposing explanation into steps, strengths and weaknesses were revealed in the studied methods, for example, on pixel clustering and perturbations in the input. Video LIME outperformed other methods with deletion values up to 31\% lower and insertion up to 30\% higher, depending on method and network. Video RISE achieved superior performance in the average drop metric, with values 10\% lower. In contrast, localization-based metrics revealed low performance across all methods, with significant variation depending on network. Pointing game accuracy reached 53\%, and IoU-based metrics remained below 20\%. Drawing on the findings across XAI methods, we further examine the limitations of the employed XAI evaluation metrics and highlight their suitability in different applications.
\end{abstract}



    % \item Decomposing removal-based explanations into steps exposes similarities across methods
    % \item REVEX exposes strengths and weaknesses of removal-based explanation methods
    % \item REVEX enabled the use of 6 image and tabular-based methods on video data
    % \item Adapted methods achieve model-agnostic explanations while maintaining good results.

\begin{highlights}
    \item Developed REVEX, a framework for removal-based video explanations in XAI
    \item Stepwise analysis reveals strengths and weaknesses of removal-based methods
    \item Adapted six XAI methods to video, integrating temporal information
    \item Video LIME excelled in deletion/insertion metrics; Video RISE in average drop metric
    \item Evaluated XAI metric limitations and their suitability for various applications
\end{highlights}

%% Keywords

\end{frontmatter}


\section{Introduction}

%As Deep Learning (DL) techniques continue to gain importance across various domains, the need for Explainable Artificial Intelligence (XAI) methods has become paramount. As Arrieta et al.\ articulated in \cite{arrieta2020explainable}, XAI endeavors not only to enhance our comprehension of ``black box'' models but also to foster user trust in their utilization, all the while upholding robust learning performance.

As deep learning (DL) gains importance across many domains, the need for explainable artificial intelligence (XAI) methods emerges. XAI endeavors to both enhance our comprehension of ``black box'' models and to foster user trust in their utilization, all while upholding robust learning performance \cite{arrieta2020explainable}.

%In recent years, numerous studies have focused on explaining DL models, especially in computer vision. Nevertheless, despite the substantial progress made in models designed for image inputs (e.g., \cite{simonyan2013deep}, \cite{bach2015pixel}, or \cite{selvaraju2017grad}), the area of explainability concerning video inputs remains relatively under-explored, posing a considerable research frontier. This scarcity of attention can be attributed to the intricacies arising from the temporal dimension in video data, which escalates the complexity of both the data itself and the models used to analyze it.

Recent studies in computer vision focus on explaining DL models. Nevertheless, despite progress with image models (e.g., \cite{simonyan2013deep,bach2015pixel,selvaraju2017grad}), explainability in video remains relatively under-explored. The scarcity of attention is likely due to the added temporal dimension in video data.  This escalates the complexity of both the data and the models to analyze them.

%Although certain studies have indeed explored video explanations, like \cite{hartley2022swag}, \cite{stergiou2019saliency}, and \cite{hiley2020explaining}, a significant portion of these attempts leans toward models specific to convolutional architectures, typically employing back-propagation techniques as in the cited works. However, our focus in this research is to investigate the application of model-agnostic explanations to the video domain, extending beyond architecture-specific solutions. While examples of model-agnostic explanation techniques for images, such as LIME \cite{ribeiro2016why}, RISE \cite{petsiuk2018rise}, or SHAP \cite{lundberg2017unified}, do exist, their direct application to the video domain presents challenges. The inclusion of the temporal dimension escalates computational demands, necessitating careful adaptation. To address this challenge, we propose the identification of common aspects in the explanation process, enabling more controlled adjustments for video data while regulating their influence on the resultant explanations. Therefore, the primary contributions of this work are threefold:

Although some studies explore video explanations, attempts often lean toward convolutional architectures, typically employing back-propagation (e.g., \cite{hartley2022swag,stergiou2019saliency,hiley2020explaining}). However, our research is directed at model-agnostic explanations that extend beyond architecture-specific solutions. Although model-agnostic explanation techniques exist for images, for example, LIME \cite{ribeiro2016why}, RISE \cite{petsiuk2018rise}, or SHAP \cite{lundberg2017unified}, their application to the video domain presents challenges. With video, the temporal dimension is added. This escalates computational demands, necessitating careful adaptation. To address this, we first identify common elements in the explanation process. This enables more controlled adjustments for video data while regulating their influence on the explanations. The primary contributions of this work are threefold:

\begin{itemize}
\item A refined, fine-grained framework for explaining model predictions tailored to computer vision, and unifying common explanation steps in existing methods.
\item An adaptation of six established explanation methods to the video domain, modifying each to accommodate the temporal dimension and local explanations.
\item A critique of the explanations generated from each method, including diverse models and datasets, to robustly assess their performance.
\end{itemize}

%This document is organized as follows: first, a comprehensive survey of related work is conducted, encompassing existing video and model-agnostic explanation methods; subsequent sections delve deep into the process of attaining model-agnostic video explanations, breaking it down into distinct well-defined stages; we then proceed to extrapolate six explanation methods to the video domain, conducting experiments with various networks and datasets; subsequently, we evaluate and compare the resultant explanations; finally, we provide concluding insights gleaned from our study. All necessary code to reproduce the experiments and to use the proposed methods has been made publicly available at: \url{https://github.com/Xavi3398/revex_framework}.

This paper is organized as follows. First, existing video and removal-based explanation methods are surveyed. Subsequent sections explore our REVEX framework for video explanations, breaking the steps into well-defined stages. We then extend six explanation methods to the video domain, conducting experiments with various networks and datasets. Subsequently, we critique the explanations. Finally, we offer insights gleaned from our research. Code to reproduce the experiments and analyses is available at \url{https://github.com/Xavi3398/revex_framework}.

\section{Related Work}

\subsection{Video Explanation Methods}

%Despite the extensive exploration of image-based explanation methods within the literature, video explanations remain comparatively under-explored. Table \ref{tab:video} provides a concise overview of studies dedicated to explaining model predictions within the context of video input. This section offers an analysis of the works presented in the table in reverse chronological order, shedding light on existing approaches and investigations tailored specifically to video data.

Despite extensive research on image-based explanation methods, video explanations are comparatively under-explored. Table \ref{tab:video} summarizes studies on video input. 

        \begin{table}[h]
        \caption{Research on Explanation Methods Using Video Data}
            \label{tab:video}
            \begin{adjustbox}{max width=\textwidth}
            \setlength\tabcolsep{4pt}
            \begin{tabular}{llll}
            \toprule
            \textbf{Ref.} & \textbf{Year} & \textbf{1st Author} & \textbf{Explanation Method}\\
            \midrule
            \cite{gaya_morey2024explainable} & 2024 & Gaya-Morey & LIME separated into spatial and temporal explanations\\
            \cite{uchiyama2023visually} & 2023 & Uchiyama  &  Adaptive occlusion sensitivity analysis (AOSA)\\
            \cite{hartley2022swag} & 2022 & Hartley & Superpixels weighted by average gradients for video (SWAG-V)\\
            \cite{li2021towards} & 2021 & Li & Spatio-temporal extremal perturbation (STEP)\\
            \cite{hiley2020explaining} & 2020 & Hiley & Selective relevance\\
            \cite{stergiou2019class} & 2019 & Stergiou &  Class feature pyramids\\
            \cite{stergiou2019saliency} & 2019 & Stergiou  & Saliency tubes\\
            \cite{hiley2019discriminating} & 2019 & Hiley  & Discriminative relevance\\
            \cite{bargal2018excitation} & 2018 & Bargal & Contrastive excitation backpropagation for RNNs (cEB-R)\\
            \cite{anders2018understanding} & 2018 & Anders &  Deep Taylor decomposition \cite{montavon2017explaining}\\
            \cite{srinivasan2017interpretable} & 2017 & Srinivasan  & Layer-wise relevance propagation (LRP) \cite{bach2015pixel}\\
            \cite{gan2015devnet} & 2015 & Gan & Class saliency maps \cite{simonyan2013deep}\\      
                \bottomrule                   
            \end{tabular}
            \end{adjustbox}
        \end{table}
        
%In \cite{gaya_morey2024explainable}, LIME was applied to video inputs by employing grid-based, separated spatial and temporal perturbations. By decomposing the video into distinct spatial and temporal dimensions and explaining each one independently, the individual contributions to the model’s predictions were assessed. This approach also demonstrated a significant acceleration in the explanation process by merging the separate explanations, as opposed to computing them across all dimensions simultaneously. However, the results greatly degrade in scenarios involving substantial temporal variations.

Gaya-Morey et al.\ \cite{gaya_morey2024explainable} applied LIME (local interpretable model-agnostic explanations) to video data by employing grid-based, separated spatial and temporal perturbations. By separately analyzing the spatial and temporal dimensions, the individual contributions to predictions were revealed. This reduced the time for explanations by merging the separate explanations, rather than computing them across all dimensions simultaneously. However, results degrade with increased temporal variation.

%Uchiyama et al. \cite{uchiyama2023visually}, extended the applicability of the Occlusion Sensitivity method \cite{zeiler2014visualizing}, originally designed for images, to videos. In the original image approach, a squared gray patch was manipulated through images to gauge its impact on predictions. In the video adaptation, they introduced optical flow as a means to extend the two-dimensional patch across the temporal dimension, aiming to trace object trajectories throughout video sequences.

Uchiyama et al.\ \cite{uchiyama2023visually} extended OSM (occlusion sensitivity method) \cite{zeiler2014visualizing} from images to videos. In the image approach, a gray patch is manipulated through an image to gauge its impact on predictions. For video, they introduced optical flow to extend the two-dimensional patch across the temporal dimension, aiming to trace object trajectories through video sequences.

%Hartley et al. introduced SWAG \cite{hartley2021swag}, a method that leverages a modified version of SLIC \cite{achanta2010slic} to compute superpixels, incorporating gradient values. The superpixels' relevance was then back-propagated to the input using guided back-propagation \cite{tobiasspringenberg2015striving}. A subsequent extension to the video domain, known as SWAG-V, incorporated the temporal dimension during superpixel computation \cite{hartley2022swag}.

Hartley et al.\ \cite{hartley2021swag} proposed   SWAG (superpixels weighted by average gradients), which extends SLIC (simple linear iterative clustering) \cite{achanta2010slic} to compute superpixels using gradients. Relevance was assessed using guided back-propagation \cite{tobiasspringenberg2015striving}. Their SWAG-V extension adds the temporal dimension (i.e., video) during superpixel assignment \cite{hartley2022swag}.

%Li et al. proposed the Spatio-Temporal Extremal Perturbation (STEP) method \cite{li2021towards}, an extension of the Extremal Perturbation (EP) technique \cite{fong2019understanding} tailored for video data. EP operats through the optimization of explanation masks, subject to distinct constraints. The extension entails modifying the EP framework to accommodate the additional temporal dimension present in videos.

Li et al.\ \cite{li2021towards} proposed STEP (spatio-temporal extremal perturbation), an extension of extremal perturbation (EP) \cite{fong2019understanding}. EP optimizes explanation masks, subject to constraints. The extension accommodates the temporal dimension in video data.

%Hiley et al. explored video explanations using deep Taylor decompositions \cite{montavon2017explaining} in \cite{hiley2019discriminating}. They disentangled spatial and temporal relevance by explaining individual frames expanded to input dimensions across various frames of the video, thus effectively removing the temporal information. The authors remarked some of the weaknesses of this approach in a later study \cite{hiley2020explaining}, where they proposed selective relevance as a better way to decompose a 3D explanation into spatial and temporal components, via derivative-based filtering to discard regions with near-constant relevance over time.

Hiley et al.\ \cite{hiley2019discriminating} explored video explanations using deep Taylor decompositions \cite{montavon2017explaining}. They disentangled spatial and temporal relevance by explaining individual frames expanded to input dimensions across video frames, thus simplifying temporal information. They identified weaknesses in this approach in a later study \cite{hiley2020explaining} and proposed selective relevance to decompose a 3D explanation into spatial and temporal components.  This was done via derivative-based filtering to discard regions with near-constant relevance over time.

%Stergiou et al. introduced two methods for explaining 3D convolutional networks \cite{stergiou2019saliency, stergiou2019class}. In \cite{stergiou2019saliency}, they presented Saliency Tubes, utilizing prediction outputs and the final activation layer to compute saliency maps, subsequently rescaled to match input dimensions. In  \cite{stergiou2019class}, they introduced Class Feature Pyramids, which employed back-propagation to traverse the network, identifying significant kernels at different depths. The resulting saliency maps yielded higher resolution compared to the Saliency Tubes approach.

Stergiou et al.\ introduced two methods for explaining 3D convolutional neural networks (CNNs). First \cite{stergiou2019saliency}, they presented saliency tubes, utilizing prediction outputs and the final activation layer to compute saliency maps rescaled to match input dimensions. Second  \cite{stergiou2019class}, they introduced class feature pyramids, which use back-propagation to traverse the network, identifying significant kernels at different depths. The resulting saliency maps yielded higher resolution than the saliency tubes approach.

%Bargal et al. focused into explaining Recurrent Neural Networks (RNNs) for video action recognition and captioning with a contrastive excitation backpropagation formulation \cite{bargal2018excitation}. This formulation marked the first instance of top-down saliency formulation within deep recurrent models, facilitating space-time grounding in video data.

Bargal et al.\ \cite{bargal2018excitation} used recurrent neural networks (RNNs) for video action recognition and captioning with a contrastive excitation back-propagation formulation. This was the first instance of top-down saliency formulation within deep recurrent models, facilitating space-time grounding in video data.

%Andersen et al. delved into video prediction explanation using the deep Taylor decomposition technique \cite{montavon2017explaining} in \cite{anders2018understanding}. Their study identified pertinent pixels and frames within the video context and addressed systematic imbalances, termed ``border'' and ``lookahead'' effects, that could occur in explanations.

Andersen et al.\ \cite{anders2018understanding} delved into video prediction explanation using deep Taylor decomposition \cite{montavon2017explaining}. They identified pertinent pixels and frames within a video and addressed systematic imbalances, termed ``border'' and ``lookahead'' effects, that occur in explanations.

%Finally, Srinivasan et al. investigated Layer-wise Relevance Propagation (LRP) \cite{bach2015pixel} for video explanations in \cite{srinivasan2017interpretable}, and Gan et al. extended in \cite{gan2015devnet} the 2D saliency maps from \cite{simonyan2013deep} to video data, yielding spatial-temporal saliency maps from 3D CNNs.

Finally, Srinivasan et al.\ \cite{srinivasan2017interpretable} investigated layer-wise relevance propagation (LRP) \cite{bach2015pixel} for video explanations, and Gan et al.\ \cite{gan2015devnet} extended Simonyan et al.'s 2D saliency maps \cite{simonyan2013deep} to video data, yielding spatial-temporal saliency maps from 3D CNNs.

\subsection{Explanations by Removal} \label{unified}

%Removal-based explanation methods, also referred to as perturbation or input-modification methods, constitute a set of techniques that share a common premise: assessing the importance of input features by analyzing how predictions change when these features are removed. This approach boasts the key advantage of model independence, as all modifications take place within the input space, with the computation of relevance relying solely on output predictions. Covert et al. conducted an extensive analysis of 26 such methods, elucidating three pivotal choices in the formulation of a removal-based explanation paradigm: the mechanism of feature removal, the model behavior for analysis, and the approach to summarizing feature influence \cite{covert2021explaining}. While some of these methods are well-established for image data, their adaptation to video entails additional considerations and choices. Given the high volume of input features, particularly in video, it is common practice to use sets of pixels as the fundamental units for explanation. Additionally, the unique characteristics of visual data necessitate adjustments, such as the replacement of pixels to simulate removal. These adaptations introduce a broader array of choices when designing the explanation process. Next, we delve deeper into six existing removal-based explanation methods that have been adapted for the video domain: LIME, SHAP, RISE, LOCO, Univariate Predictors, and Occlusion Sensitivity.

Removal-based explanation methods, also referred to as perturbation or input-modification methods, share a common purpose: assessing the importance of input features by analyzing the change in predictions when the features are removed. This approach is model-independent since modifications are in the input space, with relevance based solely on output predictions. Covert et al.\ \cite{covert2021explaining} analyzed 26 such methods and found three pivotal choices for a removal-based explanation paradigm: (i) feature removal, (ii) model behavior, and (iii) feature assessment. While some methods are well-established for image data, their adaptation to video entails additional considerations. Given the high volume of input features, particularly in video, it is common to use sets of pixels as the basis for explanation. Additionally, the unique characteristics of visual data necessitate adjustments, such as replacing pixels to simulate removal. These adaptations introduce multifaceted choices when designing an explanation method. Next, we examine six existing removal-based explanation methods that are adapted for the video domain: LIME, SHAP, RISE, LOCO, univariate predictors, and occlusion sensitivity.

%LIME \cite{ribeiro2016why}, which stands for Local Interpretable Model-agnostic Explanations, was initially proposed as a tool to elucidate predictions of diverse classifiers, although our focus is on its application to images. LIME adopts superpixels as regions for segmenting images, although it does not prescribe a specific technique. It functions by randomly occluding approximately half of the chosen regions, observing variations in class confidence, and subsequently training an interpretable surrogate model, such as linear models or decision trees, to compute the relevance of each region for different target classes. Various visualization strategies can be employed, ranging from displaying only the most significant regions to using distinct colors for positive and negative relevance.

LIME \cite{ribeiro2016why}, or local interpretable model-agnostic explanations, was intended to elucidate predictions of diverse classifiers, with an application to images. LIME adopts superpixels as regions for segmenting images, although it does not prescribe a specific technique. It functions by randomly occluding approximately half of the chosen regions, observing variations in class confidence, and subsequently training an interpretable surrogate model, such as linear models or decision trees, to compute the relevance of each region for different target classes. Various visualization strategies are employed, from displaying only the most significant regions to using distinct colors for positive and negative relevance.

%SHAP (SHapley Additive exPlanations) \cite{lundberg2017unified} introduces a unified measure of feature relevance known as Shapley values. These values express the change in expected model prediction when conditioning on a specific feature. Although multiple techniques exist for estimating Shapley values, Kernel SHAP is often favored due to its ability to yield accurate approximations with fewer passes through the model. Kernel SHAP modifies loss functions, weighting kernels, and regularization terms from the LIME framework to derive Shapley values, ensuring local accuracy and consistency violation avoidance.

SHAP (\underline{sh}apley \underline{a}dditive ex\underline{p}lanations) \cite{lundberg2017unified} introduces Shapley values, a unified measure of feature relevance. The values express the expected change in model prediction when conditioning on a specific feature. Although variations exist, Kernel SHAP is appealing as it yields accurate approximations with fewer passes. Kernel SHAP modifies loss functions, weighting kernels, and regularization terms from the LIME framework to derive Shapley values, ensuring local accuracy and avoiding consistency violation.

%Randomized Input Sampling for Explanation (RISE) \cite{petsiuk2018rise}, tailored for images, involves segmenting images with a small 2D grid that is then scaled up through bilinear interpolation to match the image dimensions. Each grid cell can either be occluded with a black color or left unaltered. A designated number of samples are generated, each with distinct combinations of perturbed regions. Region relevance is determined by averaging class predictions across samples with unperturbed regions. As RISE explanations are inherently positive, relevance is visualized using a heat map.

Randomized input sampling for explanation (RISE) \cite{petsiuk2018rise}, tailored for images, segments images in a small 2D grid that is scaled up through bilinear interpolation to match the image dimensions. Each grid is either occluded with a black color or left unaltered. A designated number of samples are generated, each with distinct combinations of perturbed regions. Region relevance is determined by averaging class predictions across samples with unperturbed regions. As RISE explanations are inherently positive, relevance is visualized using a heat map.

%Leave-One-Covariate-Out (LOCO) \cite{lei2018distribution} leverages feature ablation to ascertain the most influential feature for predictions. The model is trained iteratively, with each iteration removing a different feature from the training set to assess the resulting prediction changes. LOCO offers global explanations rather than local ones and, to the best of our knowledge, has not been applied to visual data.

Leave-one-covariate-out (LOCO) \cite{lei2018distribution} leverages feature ablation to ascertain the most influential feature. The model is trained by iteratively removing features from the training set to assess the change in prediction. LOCO offers global explanations rather than local ones and, to the best of our knowledge, has not been applied to visual data. 

%Univariate Predictors \cite{guyon2003introduction} are advocated over multivariate counterparts for reasons encompassing potential performance enhancements, computational efficiency, and enhanced interpretability. This method predominantly explores various variable selection strategies based on expected loss across the dataset. Similarly to LOCO, Univariate Predictors provide global rather than local explanations and have not yet been employed in the context of visual data, as far as we are aware.

Univariate predictors \cite{guyon2003introduction} are advocated over multivariate counterparts for  performance enhancements, computational efficiency, and interpretability. This method tests variable selection strategies on expected loss across the dataset. Similarly to LOCO, univariate predictors provide global rather than local explanations. They are not yet used with visual data, as far as we know.

%Occlusion Sensitivity \cite{zeiler2014visualizing} finds pixel relevance by sliding a gray patch across the image and observing the corresponding model prediction changes for a specific class. Since each pixel corresponds to a distinct occlusion, the pixel's relevance aligns directly with the model prediction. This approach yields a heat map representation of relevance, where colder areas signify regions of utmost significance due to the greatest decrease in class confidence. While this method has been extended to the video domain using optical flow in \cite{uchiyama2023visually}, a more natural extension can be achieved by simply adding an additional dimension to the occlusion kernel. This approach offers advantages over optical flow, including better performance on videos featuring cuts, camera movements, and objects entering or exiting the camera's field of view.

Occlusion sensitivity \cite{zeiler2014visualizing} finds pixel relevance by sliding a gray patch across an image and assessing the change in prediction. Since each pixel is an occlusion, the pixel's relevance correlates with the model prediction. The result is a heat map for relevance, where colder areas are regions of higher significance due to a decrease in class confidence. While this method extends to video using optical flow \cite{uchiyama2023visually}, a natural extension adds an additional dimension to the occlusion kernel. This offers advantages over optical flow, including better performance on videos containing cuts, camera movements, or objects entering or leaving the field of view.

\section{The REVEX Framework: Removal-based Video Explanations}

%As commented in the preceding section, Covert et al. \cite{covert2021explaining} introduced a comprehensive framework for removal-based explanations encompassing three key decisions: the manner in which features are removed, the analysis of model behavior, and the summary of feature influence. While some of the reviewed explanation techniques can be applied to image inputs, additional considerations are necessary, and these become even more significant when extending to the temporal dimension of videos. Hence, prior to analyzing the video-specific explanation methods, we expand upon Covert et al.'s framework by incorporating additional choices tailored for visual data.
    
As noted above, Covert et al.\ \cite{covert2021explaining} introduced a framework for removal-based explanations focusing on feature removal, model behavior, and feature assessment.  While some of the techniques are for images, additional considerations are important in the temporal dimension of videos. Hence, prior to analyzing video-specific explanation methods, we expand upon Covert et al.'s framework by incorporating additional choices for visual data.
    
%More specifically, we introduce the segmentation choice prior to Covert et al.'s feature removal (their first choice). We then subdivide their feature removal choice into three distinct aspects: feature selection, sample selection, and feature removal. We retain their choices for model behavior and summary techniques unchanged. Lastly, we append the visualization choice at the conclusion of the framework. A comparison between the two frameworks is presented in Figure \ref{fig:frameworks}. In the subsequent subsections, we address each of these choices individually, scrutinizing the principal advantages and limitations of the considered options. We omit the exploration of model behavior and summary techniques, as they have already been addressed in \cite{covert2021explaining}.

\begin{figure}[ht]
         \centering
         \includegraphics[width=0.6\linewidth]{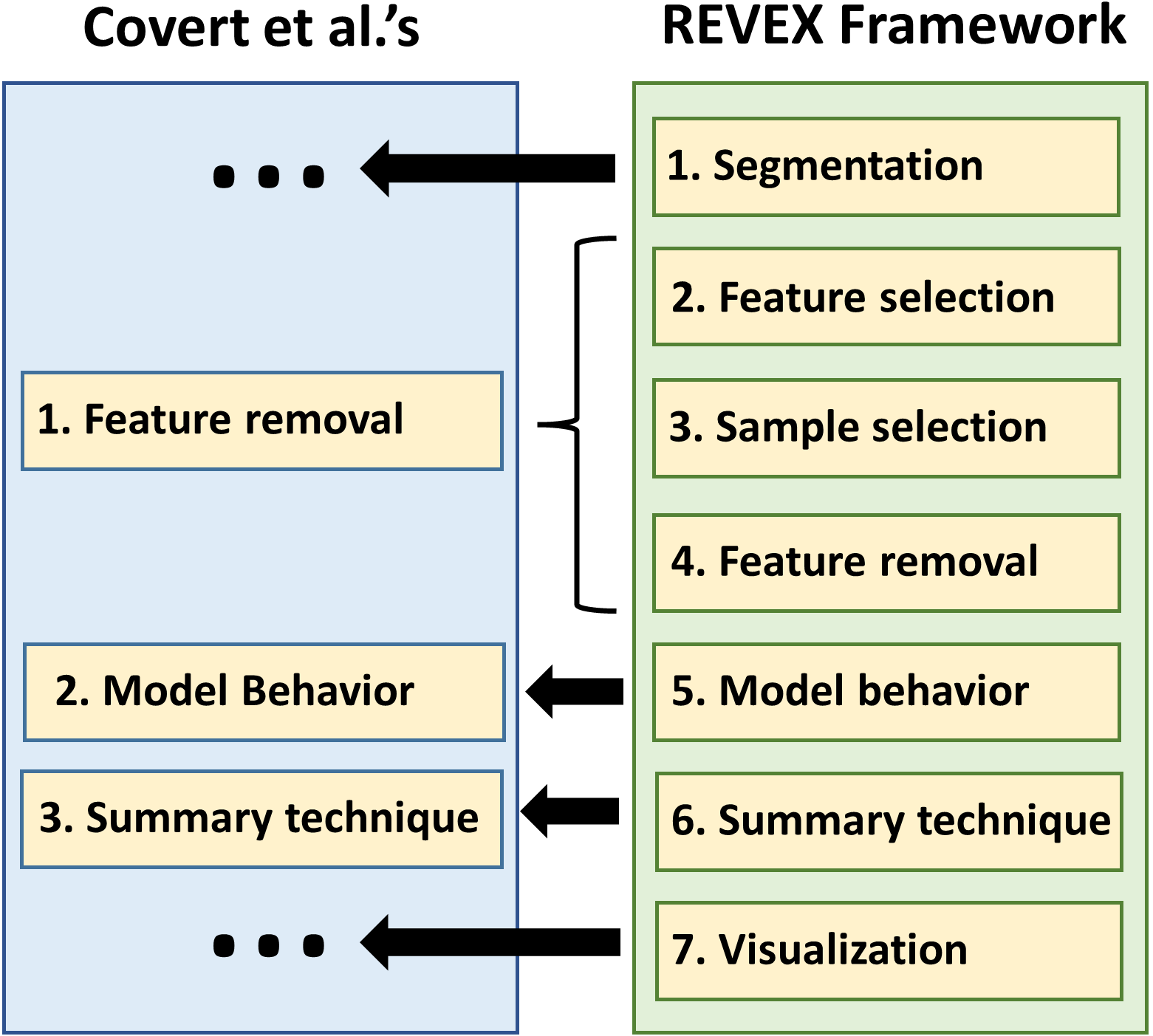}
         \caption{Covert et al.'s framework \cite{covert2021explaining} vs. REVEX, our extended framework. Our proposal begins and ends with segmentation and visualization, while decomposing feature removal into three  components.}
        \label{fig:frameworks}
    \end{figure}

More specifically, we insert a segmentation choice prior to Covert et al.'s feature removal (their first choice). We then divide their feature removal into three parts: feature selection, sample selection, and feature removal. We retain unchanged their choices for model behavior and summary techniques. Lastly, we append the visualization choice at the conclusion of the framework. A comparison of the two frameworks is presented in Figure \ref{fig:frameworks}, with our proposal henceforth referred to as the REmoval-based Video EXplanations (REVEX) Framework. In subsequent sections, we address each choice individually, scrutinizing advantages and limitations. We omit exploring model behavior and summary techniques, as they are addressed by Covert et al.\ \cite{covert2021explaining}.

\subsection{Segmentation}

%An integral challenge encountered when working with visual explanations is the sheer volume of data. Explanation through removal operates at the level of features, which often makes operating at the pixel and frame levels infeasible. Thus, a common approach is to segment input images or videos into regions, comprising collections of pixels or pixels from diverse frames. This step significantly influences the resultant explanations, as pixels grouped within the same region are treated as a single feature, inevitably assigning them identical relevance. Figure \ref{fig:seg_video} depicts diverse strategies for video segmentation.

\begin{figure}[h]
             \centering
             \includegraphics[width=.8\textwidth]{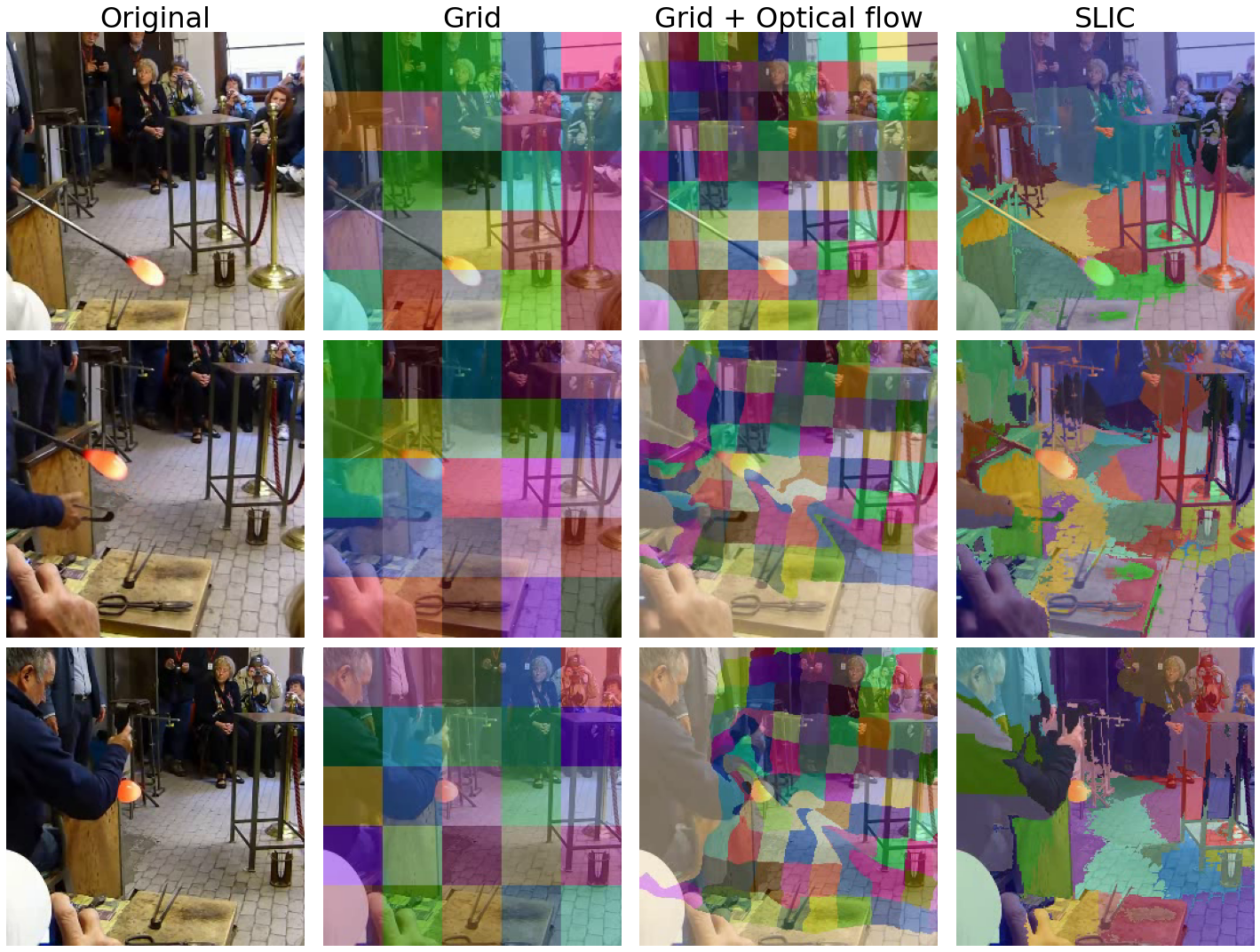}
             \caption{Video segmentation. From top to bottom, initial, middle, and final frames. From left to right, unmodified video, 3D grid, 2D grid segmentation extended with optical flow, and SLIC. Optical flow estimation was performed using PWC-Net \cite{sun2018pwc}.}
            \label{fig:seg_video}
        \end{figure}

A challenge encountered with visual explanations is the sheer volume of data. Explanation through removal operates at the level of features, which often makes operating at the pixel and frame levels infeasible. Thus, a common approach is to segment input images or videos into regions, comprising collections of pixels or pixels from diverse frames. This step significantly influences the explanations, as pixels grouped within the same region are assigned identical relevance. Figure~\ref{fig:seg_video} depicts strategies for video segmentation.

%One straightforward method for segmenting images or videos involves employing a 2D or 3D grid to establish rectangular or cuboidal patches. While this method boasts simplicity and expedited region generation, it overlooks pixel color information, potentially resulting in similar and proximal pixels being distributed across different regions. This approach is employed in RISE explanations \cite{petsiuk2018rise} for images and can be extended to videos by adding the temporal dimension to the grid, as depicted in Figure \ref{fig:seg_video}.

One way to segment images or videos is through a 2D or 3D grid of cuboidal patches. While simple, it overlooks pixel color and distributes similar and proximal pixels across different regions. An example is RISE explanations \cite{petsiuk2018rise} which extend to videos by adding the temporal dimension to the grid. See Figure \ref{fig:seg_video}.

%Alternatively, the adoption of superpixels as features is another option, wherein pixels are grouped based on similarity metrics. Unlike the grid-based approach mentioned earlier, this method is more intricate and time-consuming, yet it gathers comparable pixels into the same regions. For videos, pixels from different frames can be incorporated during superpixel computation. Various superpixel segmentation options for images are demonstrated in Figure \ref{fig:seg_image}, encompassing Simple Linear Iterative Clustering (SLIC) \cite{achanta2010slic}, which employs K-means in a 5D color space; Quick Shift \cite{vedaldi2008quick}, a kernelized mean-shift approximation; Compact Watershed \cite{neubert2014compact}, utilizing marker-driven flooding to define regions with control over compactness; and Felzenswalb's efficient graph-based image segmentation \cite{felzenszwalb2004efficient}. The number of segments and their compactness (the balance between spatial and color distance) can be governed by diverse hyperparameters contingent upon the method. Implementation of these techniques can be found in scikit-image's segmentation package\footnote{\url{https://scikit-image.org/docs/stable/api/skimage.segmentation.html}}, although only the SLIC segmentation is applicable to video data. Last column in Figure \ref{fig:seg_video} illustrates an example of SLIC segmentation applied to video.

\begin{figure}[h]
             \centering
             \includegraphics[width=\textwidth]{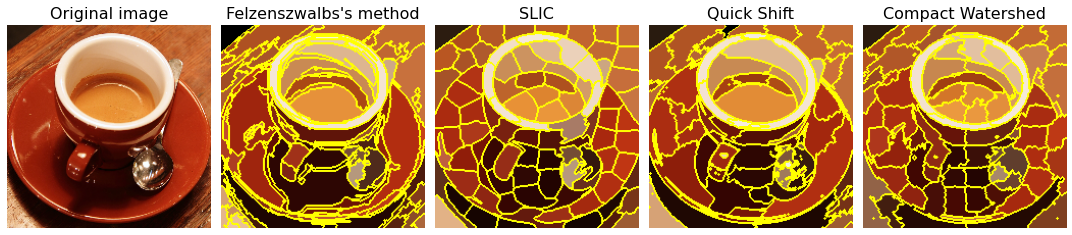}
             \caption{Example image (left) and image segmented using Felzenszwalbs's method, SLIC, quick shift, and compact watershed.  Segmented regions are outlined in yellow and filled with the average color for each region.}
            \label{fig:seg_image}
        \end{figure}

Another option uses superpixels as features, wherein pixels are grouped based on similarity. Unlike grids, mentioned earlier, this method is more intricate and time-consuming, yet it gathers comparable pixels into the same regions. For videos, pixels from different frames are incorporated during superpixel computation. Figure~\ref{fig:seg_image} shows examples of superpixel segmentation for images, including Felzenszwalb's efficient graph-based image segmentation \cite{felzenszwalb2004efficient},  simple linear iterative clustering (SLIC) using K-means in a 5D color space \cite{achanta2010slic}, quick shift using a kernelized mean-shift approximation \cite{vedaldi2008quick}, and compact watershed utilizing marker-driven flooding to define regions with control over compactness \cite{neubert2014compact}. The number of segments and their compactness (the balance between spatial and color distance) is governed by hyperparameters contingent on the method. Implementation of these techniques is found in scikit-image's segmentation package\footnote{\url{https://scikit-image.org/docs/stable/api/skimage.segmentation.html}}, although only the SLIC segmentation is for video data. The last column in Figure~\ref{fig:seg_video} is an example of SLIC segmentation applied to video.

%Finally, while many image segmentation methods can be extrapolated to video, an additional approach involves computing optical flow within the video and utilizing it to expand image segmentation from the initial frame to subsequent ones, as demonstrated in \cite{uchiyama2023visually} for occlusion explanations in videos. However, potential complications arise, particularly when segmenting objects that do not appear in the first frame. This issue is evident in Figure \ref{fig:seg_video}, where a substantial region encompassing the individual on the left emerges towards the video's conclusion, a result of the individual not appearing in the initial frame. Moreover, this type of segmentation generates segments that span the entire temporal frame, which inhibits the ability to provide temporal explanations, as all frames are assigned the same relevance.        
        
Finally, while many image segmentation methods extend to video, optical flow works by extending the segmentation from the first frame to the whole video by following video motion, as demonstrated by Uchiyama et al.\ \cite{uchiyama2023visually} for occlusion explanations in videos. However, complications arise when segmenting objects absent in the first frame. This is evident in Figure~\ref{fig:seg_video} where a substantial region encompassing the individual on the left emerges towards the video's conclusion, a result of the individual not appearing in the initial frame. Moreover, this generates segments that span the entire temporal frame, which degrades temporal explanations, as all frames are assigned the same relevance.        
        
\subsection{Feature Selection}

%Feature removal methods operate by observing the impact on the model's predictions upon removing (or occluding) specific features (or regions). However, certain features may exhibit a more pronounced influence when removed collectively with others, prompting many explanation methods to employ simultaneous removal of multiple features. Given that the number of features removed per sample can influence explanation outcomes, this choice warrants careful consideration during the establishment of the explanation process. Various possibilities for this selection are illustrated in Figure \ref{fig:feature_selection}.
        
\begin{figure}[h]
             \centering
             \includegraphics[width=0.6\linewidth]{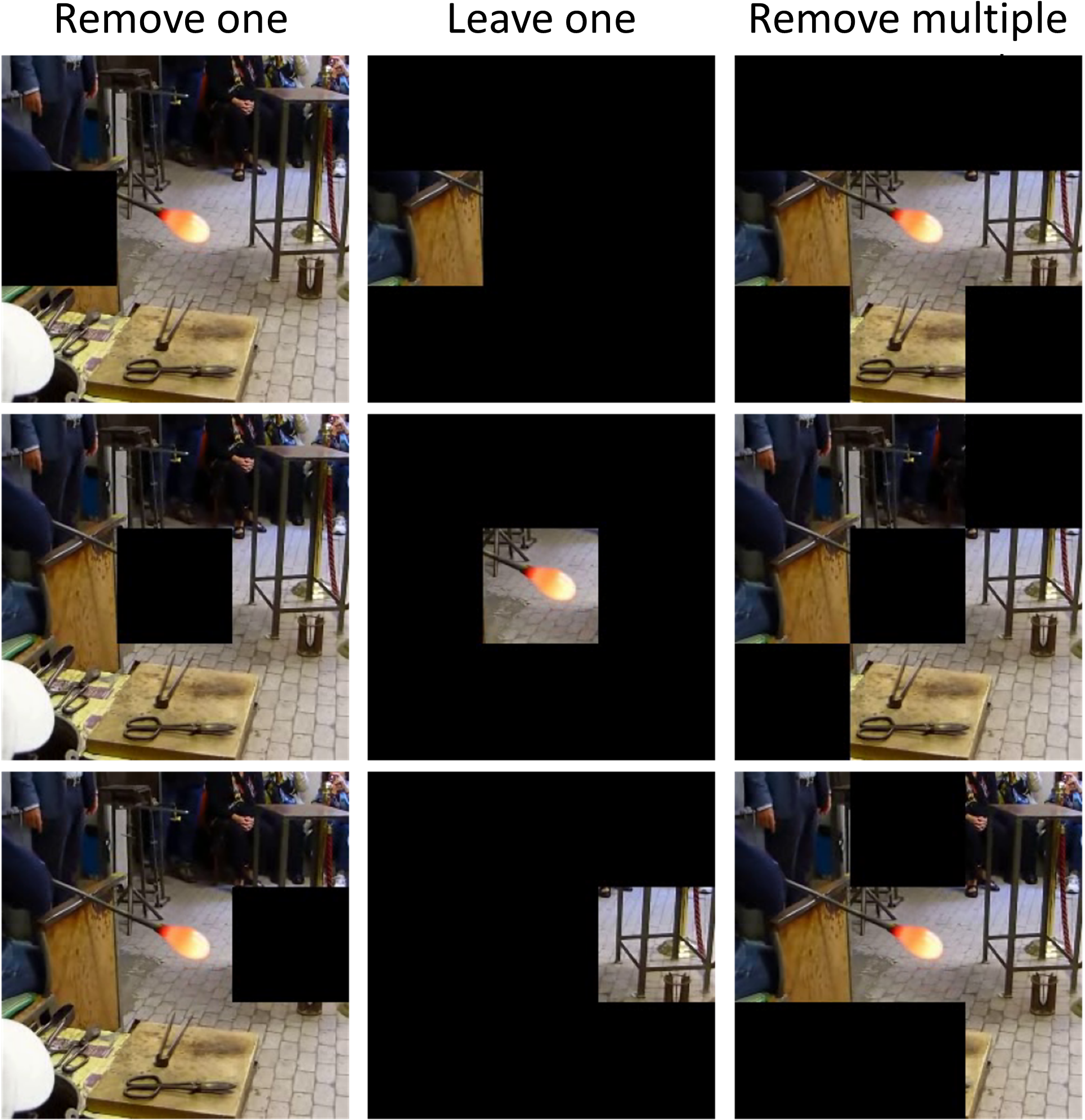}
             \caption{Feature selection steps. Individual frames are displayed, although the regions span subsequent frames. Left to right, samples with a single feature removed, with all but one removed, and with about half the regions removed. For visualization, 3D grid-based segmentation and black fill were employed to remove regions.}
            \label{fig:feature_selection}
        \end{figure}

Feature removal operates by observing the impact on the model's predictions after removing (or occluding) features (or regions). However, certain features may exhibit a more pronounced influence when removed collectively, prompting many methods to simultaneously remove multiple features. Given that the number of features removed per sample influences outcomes, the choice warrants consideration. Possibilities for feature selection are illustrated in Figure~\ref{fig:feature_selection}.
        
%One straightforward approach is to remove a single feature for each sample, enabling the assessment of individual feature relevance in model decisions. However, this approach lacks the ability to reveal complex relationships in relevance that become apparent when multiple features are removed simultaneously. Methods employing this form of feature removal include Occlusion Sensitivity \cite{zeiler2014visualizing} and its adaptation to video \cite{uchiyama2023visually}, as well as LOCO \cite{lei2018distribution}. Conversely, an alternative approach involves removing all features except one, as exemplified by the Univariate Predictors method \cite{guyon2003introduction}.

A simple approach is removing one feature per sample. This enables assessing single feature relevance in model decisions, but downplays the complex relationships seen when multiple features are removed simultaneously. Methods employing this technique include occlusion sensitivity \cite{zeiler2014visualizing} and one of its adaptations to video, AOSA-SGL \cite{uchiyama2023visually}, as well as LOCO \cite{lei2018distribution}. An alternative is to remove all but one feature, as with univariate predictors \cite{guyon2003introduction}.

%Alternatively, multiple features can be removed simultaneously, facilitating the identification of significant feature sets. However, this poses a more intricate challenge in uncovering feature relevance. Approaches such as LIME \cite{ribeiro2016why}, RISE \cite{petsiuk2018rise}, and Kernel SHAP \cite{lundberg2017unified} adopt this strategy. In LIME and RISE, each region is either removed or retained with a probability of $p=0.5$, resulting in the occlusion of approximately half the regions. This choice optimizes the minimization of sample requirements for an accurate approximation of feature relevance. It is worth noting that altering the probability value to regulate the extent of occluded regions is feasible, and even a hybrid of samples with varying numbers of regions could be employed, although this would likely necessitate an increase in the number of perturbed samples. Kernel SHAP, for instance, follows this strategy by employing a mix of samples with either a few or a greater number of occluded features than those with an approximately uniform amount of occlusions.

Alternatively, removing multiple features simultaneously can identify significant feature sets. However, this poses challenges in uncovering feature relevance. LIME \cite{ribeiro2016why}, RISE \cite{petsiuk2018rise}, and Kernel SHAP \cite{lundberg2017unified} adopt this strategy. In LIME and RISE, each region is removed or retained with probability $p = .5$, thereby occluding about half the regions. This minimizes the sample requirements for an accurate approximation of feature relevance. It is of course possible to alter $p$ to regulate the number of occluded regions or adopt a hybrid method, although this may increase the number of perturbed samples. Kernel SHAP, for instance, follows this strategy by employing a mix of samples with different numbers of occluded features than with uniform occlusions.

\subsection{Sample Selection}
    
%To effectively discern the significance of various features, removal-based methods necessitate perturbed samples of the input to correlate model prediction differences with features. Each explanation requires a specific number of model passes, corresponding to the number of samples used, which may vary among explanation methods. Often treated as a hyperparameter, the number of samples can substantially impact the quality of explanation estimates. Achieving a balance between estimation accuracy and computational efficiency is vital; it should be sufficiently high to yield insightful explanations while minimized to reduce the computational load.

To determine the significance of features, removal-based methods use perturbed input samples to correlate model prediction differences with each feature. Each explanation requires a specific number of model passes, corresponding to the number of samples, which may vary among explanation methods. Often treated as a hyperparameter, the number of samples impacts the quality of explanation. Achieving a balance between estimation accuracy and computational efficiency is vital: The explanation should be insightful yet not burdensome on the computational load.

%Certain methods mandate a predetermined number of samples, such as the previously mentioned Occlusion Sensitivity \cite{zeiler2014visualizing}, which depends on pixel count and occlusion kernel size, as well as LOCO \cite{lei2018distribution} and Univariate Predictors \cite{guyon2003introduction}, both reliant on feature count. Conversely, methods like LIME \cite{ribeiro2016why}, RISE \cite{petsiuk2018rise}, or Kernel SHAP \cite{lundberg2017unified} require a user-defined sample count. Generally, increasing the sample count augments the quality of approximated relevance. This is evident in Shapley value estimation using Kernel SHAP, where testing all possible feature combinations is unfeasible; however, a close approximation can be achieved through a substantial sample count, leading to more accurate Shapley value estimates. The appropriate number of samples depends significantly on the feature count, hence video explanations generally necessitate more samples than image explanations, unless the number of regions is curtailed by enlarging their size. In the RISE publication's experiment \cite{petsiuk2018rise}, 4,000 or 8,000 samples were used to compute explanations, contingent on the model; the public LIME image implementation defaults to 1,000 samples; and in the Kernel SHAP implementation, the default count is 2,048 plus the number of features.

Certain methods, such as occlusion sensitivity \cite{zeiler2014visualizing}, predetermine the number of samples based on pixel count and occlusion kernel size.  This is also the case with LOCO \cite{lei2018distribution} and univariate predictors \cite{guyon2003introduction} which rely on feature count. Conversely, LIME \cite{ribeiro2016why}, RISE \cite{petsiuk2018rise}, or Kernel SHAP \cite{lundberg2017unified} require a user-defined sample count. Generally, increasing the sample count improves the relevance quality. This is evident using Kernel SHAP, where testing all feature combinations is unfeasible; however, an approximation is achieved through a substantial sample count, leading to more accurate shapley value estimates. The appropriate number of samples depends on the feature count, hence video explanations generally require more samples than image explanations, unless the number of regions is reduced by enlarging their size. In the RISE experiment \cite{petsiuk2018rise}, 4,000 or 8,000 samples were used to compute explanations, contingent on the model; the public LIME image implementation defaults to 1,000 samples; and in the Kernel SHAP implementation, the default is 2,048 plus the number of features.

\subsection{Feature Removal}

%In the context of visual data, the concept of removing a pixel does not directly apply; instead, we substitute (or occlude) it with a designated color, effectively nullifying the information it carries. This section delves into various approaches for performing this occlusion, as demonstrated in Figure \ref{fig:feature_removal}.
        
\begin{figure}[h]
             \centering
             \includegraphics[width=0.75\linewidth]{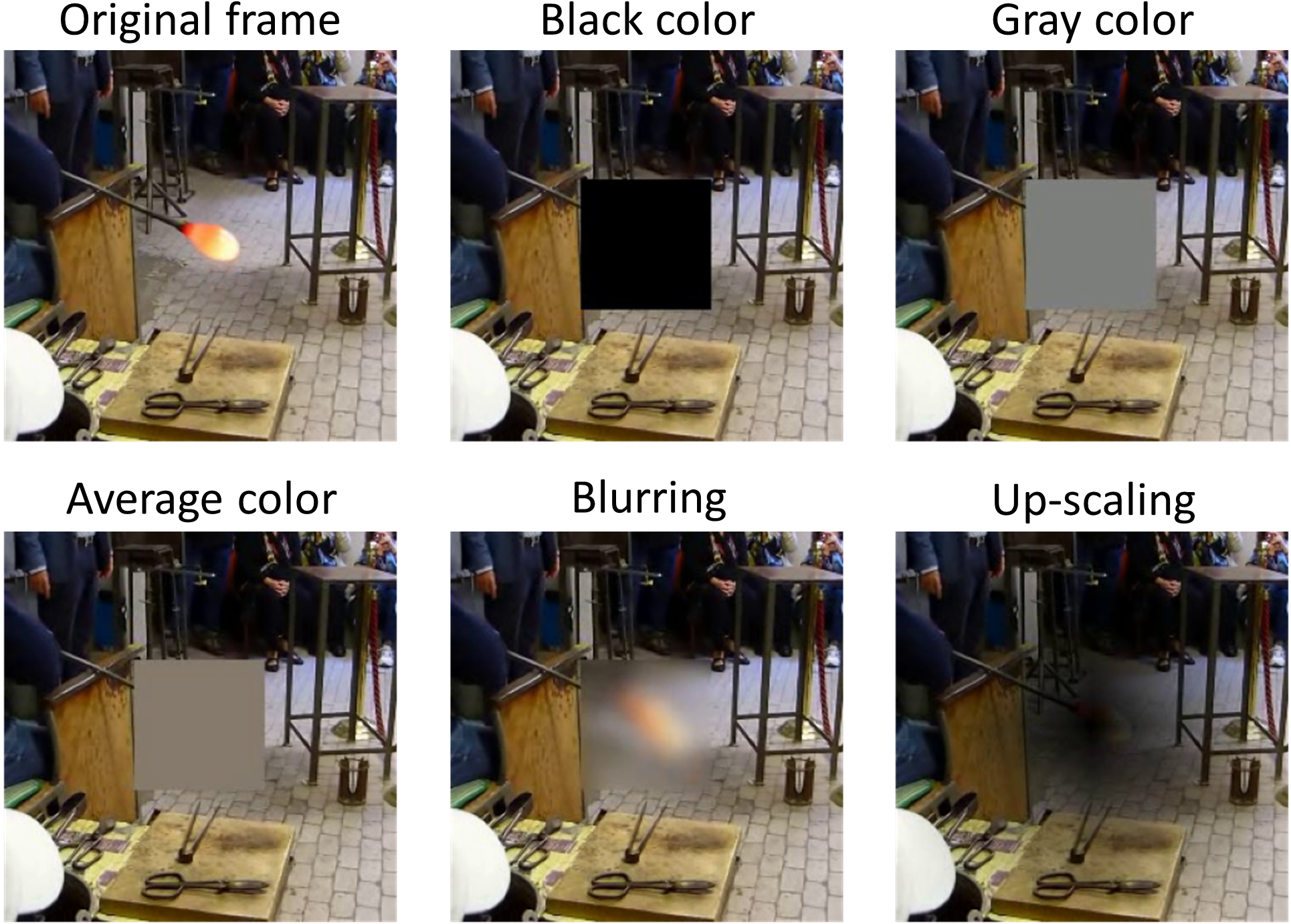}
             \caption{Feature removal methods. As labeled, original input, occlusion with black, gray, average color, uniform blur filter, and up-scaled black. Only a single frame is displayed, however, segmentation uses a 3D grid (i.e., across frames). For clarity only one region is occluded in each sample.}
            \label{fig:feature_removal}
        \end{figure}

For visual data, removing a pixel does not directly apply.  Instead, we substitute (or occlude) the pixel with a color, effectively nullifying the information it carries. This section examines approaches for performing this occlusion, as demonstrated in Figure~\ref{fig:feature_removal}.
        
%From a color perspective, a straightforward method for occlusion involves filling all regions with a consistent color. Common choices include black (e.g. RISE \cite{petsiuk2018rise}) or gray (e.g. Occlusion Sensitivity \cite{zeiler2014visualizing}). However, these approaches might suffer from the drawback of generating out-of-domain data by introducing colors not present in the training dataset \cite{miro-nicolau2024assessing}. To mitigate this concern, a solution is to adopt distinct colors based on the occluded region. For example, the mean color of the occluded region can be used as a replacement, as employed by the image implementation of LIME \cite{ribeiro2016why}. Another technique is featured in \cite{fong2017interpretable}, wherein occlusion is achieved through blurring the region in the image. Alternatively, more sophisticated methods, such as image inpainting \cite{xie2012image}, could be employed to replace the region in a more intelligent manner. The underlying idea is to eliminate the necessary information for class recognition while preserving the colors present in the image or video. Moreover, it's possible to retain some level of transparency within the used mask rather than entirely removing the region. This approach is seen in RISE explanations \cite{petsiuk2018rise}, where a low-resolution grid is initially constructed and then scaled up to produce the transparency gradient. This technique allows for smoother occlusion, avoiding the introduction of sharp edges that could result from abrupt occlusion.
    
A simple method for occlusion is to fill regions with a color, for example, black (RISE \cite{petsiuk2018rise}) or gray (occlusion sensitivity \cite{zeiler2014visualizing}). Since the result might include out-of-domain data, that is, colors not in the training dataset \cite{miro-nicolau2024assessing}, an alternate method is to use the mean color of the occluded region, as with LIME \cite{ribeiro2016why}. Another technique is blurring, as described by Fong and Vedaldi \cite{fong2017interpretable}. More sophisticated methods, such as image inpainting \cite{xie2012image}, replace regions in a more intelligent manner such as eliminating information for class recognition while preserving colors in the image or video. Moreover, it is possible to use transparency (i.e., alpha) within the mask rather than entirely removing a region. An example is RISE \cite{petsiuk2018rise}, where a low-resolution grid is constructed and scaled up to produce the transparency gradient. This technique creates smoother occlusion and avoids sharp edges.

\subsection{Visualization}

%The final step in the explanation process is visualization, and although it might appear straightforward, it significantly impacts user comprehension of the explanations. Visualization methods vary based on the explanation technique employed, and this section delves into the decisions to be made in this regard.

The final step in the explanation process is visualization.  Although, it might appear straightforward, visualization significantly impacts user comprehension of the explanations \cite{hudon2021explainable}. Visualization methods vary based on the explanation technique employed, as explored in this section.

%An essential divergence among explanation techniques lies in whether computed relevance can solely be positive or whether it can also extend into negative values. Positive relevance is attributed to regions whose removal diminishes confidence in a particular class, indicating their necessity for accurate class prediction. Conversely, negative relevance pertains to regions that, when removed, increase confidence in a class, which means that they mislead the model toward incorrect predictions when present. Methods like LIME \cite{ribeiro2016why} and SHAP \cite{lundberg2017unified} compute both positive and negative relevance, while RISE \cite{petsiuk2018rise} and Occlusion Sensitivity \cite{zeiler2014visualizing} exclusively produce positive relevance. Visualization of positive relevance often involves a heat map, representing the region's importance through variations in color. When incorporating both positive and negative relevance, distinct shades of two colors may be used to differentiate them. For instance, LIME uses green and red to denote positive and negative relevance, respectively. An alternative approach involves focusing solely on positive or negative relevance, allowing for the use of heat maps for visualization.

A divergence among explanation techniques is whether relevance is solely positive or can also include negative values. Positive relevance occurs when removing a region diminishes confidence in a class, indicating the class's necessity for accurate class prediction. Conversely, negative relevance occurs when removing a region increases confidence in a class, which means that the class mislead the model when present. Methods like LIME \cite{ribeiro2016why} and SHAP \cite{lundberg2017unified} include both positive and negative relevance, while RISE \cite{petsiuk2018rise} and occlusion sensitivity \cite{zeiler2014visualizing} only produce positive relevance. Visualizing positive relevance often uses a heat map with color representing the region's importance. When incorporating both positive and negative relevance, distinct colors may differentiate them. For instance, LIME uses green and red for positive and negative relevance, respectively. An alternative approach focuses individually on positive or negative relevance and uses heat maps for visualization.

%To enhance visualization, histogram stretching can be applied to re-scale relevance values within the available color space, aiding in the differentiation of relevance across regions. Additionally, rather than displaying all regions, a subset can be chosen, be it the top $n$ significant regions, those surpassing a minimum relevance threshold, or those starting from the most important and incrementally accumulating relevance until a set threshold is reached. These methods of reducing displayed regions simplify and refine the explanations by mitigating noise.

To enhance visualization, histogram stretching can be applied to re-scale relevance within a color space, aiding in the differentiation of relevance across regions. Additionally, rather than displaying all regions, a subset can be chosen, such as the $n$ most relevant regions, regions surpassing a minimum relevance threshold, or regions starting from the most important and incrementally accumulating relevance until a set threshold is reached. These methods simplify and refine the explanations by mitigating noise.

%Furthermore, explanations can be merged with the input image or video, affording a direct view of the relevance of each pixel for prediction. This can be achieved through alpha compositing, blending explanations with the input data while specifying a transparency value, allowing representation of explanation magnitude. Alternatively, when magnitude is less significant, specific pixel values of the $n$ most relevant regions can be altered. However, care must be taken to consider the input data's influence on interpretation, as it can introduce color variations potentially leading to incorrect associations with relevance. For instance, if relevance is displayed in green and the scene already includes green elements, the merge might become ambiguous. This can be addressed by adjusting chosen colors for relevance display or by applying higher transparency to input data and lower transparency to explanations, accentuating explanation colors over the background.

Explanations can merge with the input image or video, affording a direct view of the relevance of each pixel. This is possible through alpha compositing; that is, blending explanations with the input data using transparency as a weighting. Alternatively, when magnitude is less significant, specific pixels in the $n$ most relevant regions can be altered. However, care is warranted in considering the influence of input data on interpretation, as this can lead to color variations that incorrectly imply relevance. For instance, if relevance is displayed in green and the scene includes green elements, the merge might be ambiguous. This is addressed by adjusting  colors for relevance display or by applying higher transparency to input data and lower transparency to explanations, accentuating explanation colors over the background.

        Table \ref{tab:choices} summarizes options to consider at each stage of the explanation process. Note that for certain steps, such as visualization, the options are not mutually exclusive.

        \begin{table}[H]
        \caption{Options for Each Explanation Stage}
        \label{tab:choices}
        \begin{tabularx}{\textwidth}{X|X}
        
            \toprule
            \textbf{1. Segmentation} & \textbf{2. Feature selection} \\
            \midrule
            \begin{itemize}[leftmargin=1.2em]
                \setlength\itemsep{-.3em}
                \vspace{-21pt}
                \item 3D grid
                \item 3D superpixels
                \item 2D segmentation with optical flow
                \vspace{-12pt}
            \end{itemize} & 
            \begin{itemize}[leftmargin=1.2em]
                \setlength\itemsep{-.3em}
                \vspace{-21pt}
                \item Removal of $n$ regions
                \item Remove with probability $p$
                \item Fixed number of features removed
                \item Removal of features based on a distribution
                \vspace{-12pt}
            \end{itemize} \\
            
            \toprule
            \textbf{3. Sample selection} & \textbf{4. Feature removal} \\
            \midrule
            \begin{itemize}[leftmargin=1.2em]
                \setlength\itemsep{-.3em}
                \vspace{-21pt}
                \item Fixed value
                \item Dependent on the number of features
                \vspace{-12pt}
            \end{itemize} &
            \begin{itemize}[leftmargin=1.2em]
                \setlength\itemsep{-.3em}
                \vspace{-21pt}
                \item Fixed color
                \item Mean color
                \item Mean color of the removed region
                \item Blurring
                \item Upscaling
                \item Inpainting
                \vspace{-12pt}
            \end{itemize} \\
            
            \toprule
            \textbf{5. Model behavior} & \textbf{6. Summary technique} \\
            \midrule
            \begin{itemize}[leftmargin=1.2em]
                \setlength\itemsep{-.3em}
                \vspace{-21pt}
                \item Prediction
                \item Prediction loss
                \item Dataset loss
                \vspace{-12pt}
            \end{itemize} & 
            \begin{itemize}[leftmargin=1.2em]
                \setlength\itemsep{-.3em}
                \vspace{-21pt}
                \item Mean confidence of feature when included
                \item Difference unperturbed and perturbed features
                \item Shapley value
                \item Linear model
                \vspace{-12pt}
            \end{itemize} \\ 
            
            \toprule
            \multicolumn{2}{l}{\textbf{7. Visualization}} \\
            \midrule
            \multicolumn{2}{l}{ \tabitem  Heatmap to visualize magnitudes} \\
            \multicolumn{2}{l}{ \tabitem  Preserve/remove negative relevance} \\
            \multicolumn{2}{l}{ \tabitem  Display only the $n$ most important regions} \\
            \multicolumn{2}{l}{ \tabitem  Minimum relevance threshold} \\
            \multicolumn{2}{l}{ \tabitem  Maximum accumulated relevance threshold} \\
            \multicolumn{2}{l}{ \tabitem  Histogram stretching} \\
            \multicolumn{2}{l}{ \tabitem  Alpha composing} \\
            \bottomrule
        \end{tabularx}
        \end{table}

\section{Methodology}

%Following the described framework for conducting local agnostic explanations, as outlined in the previous section, we extend well-established removal-based explanation methods to accommodate video data. To achieve this, we initially trained three distinct models that form the focal point of our explanations. These models were trained on three datasets characterized by stark dissimilarities. By employing diverse networks and datasets, our intention was to introduce a substantial range in predictions to be explained. This approach allows for a comprehensive evaluation of the explanation methods across various models, environmental contexts, and classes.

Following our REVEX framework (see Figure~\ref{fig:frameworks}), we extend removal-based explanation methods to video data. Three models form the basis for our explanations. These were trained on three dissimilar datasets to yield a range in predictions. Thus our evaluation spans different models, environmental contexts, and classes.

\subsection{Networks}
\label{sec:networks}

%We chose three distinct networks to form the foundation of our experiment: TimeSformer \cite{timesformer}, TANet \cite{tanet}, and TPN \cite{tpn}. These selections were motivated by their robust performance in action classification tasks and their user-friendly implementation within MMAction2 \cite{2020mmaction2}, an open-source PyTorch-based toolbox designed for video comprehension. Their basic specifications can be seen in Table \ref{tab:models}.

The networks used in our experiment are TimeSformer \cite{timesformer}, TANet \cite{tanet}, and TPN \cite{tpn}. These were chosen due to their robust performance in action classification tasks and their user-friendly implementation within MMAction2 \cite{2020mmaction2}, an open-source PyTorch-based toolbox for video comprehension. Their specifications are summarized in Table~\ref{tab:models}. The top-1 and top-5 accuracy metrics are from MMAction2 \cite{2020mmaction2} and were computed on the Kinetics 400 dataset with pre-training on ImageNet.

        \begin{table}[h]
        \caption{Network Specifications and Performance Benchmarks}
        \label{tab:models}
        \setlength\tabcolsep{3.5pt}
        \begin{adjustbox}{max width=\textwidth}
        \begin{tabular}{l|llll|ll}
            \toprule
            & & & & & \multicolumn{2}{c}{\textbf{Accuracy}} \\
            \textbf{Network}       & \textbf{Backbone}            & \textbf{Resolution} & \textbf{Frames} & \textbf{Crops} & \textbf{top-1~~} & \textbf{top-5} \\
            \midrule
            \textbf{TimeSformer} & TimeSformer (divST) & 224 $\times$ 224    & 8      & 3     & 77.69    & 93.45    \\
            \textbf{TPN}         & ResNet50            & 224 $\times$ 224    & 80     & 3     & 76.74    & 92.57    \\
            \textbf{TANet}       & ResNet50            & 224 $\times$ 224    & 80     & 3     & 76.25    & 92.41 \\
            \bottomrule
        \end{tabular}
        \end{adjustbox}
        \end{table}

%TimeSformer is an adaptation of the Vision Transformer architecture, designed to incorporate spatio-temporal feature learning by leveraging frame-level patches. Among its various iterations, we have chosen the ``divided attention'' version, which demonstrated optimal performance. This version excels by independently computing spatial and temporal attention.

TimeSformer adapts the vision transformer architecture \cite{dosovitskiy2021image} to incorporate spatio-temporal feature learning via frame-level patches. We used the ``divided attention'' version, which demonstrates optimal performance by independently computing spatial and temporal attention.

%TANet encompasses a Temporal Adaptive Module (TAM) embedded within a 2D CNN architecture. This configuration introduces minimal computational overhead while effectively capturing both short and long-term information through a two-level adaptive scheme. The TAM incorporates a local temporal window and a globally derived aggregation weight to achieve this.

TANet encompasses a temporal adaptive module (TAM) embedded in a 2D CNN architecture. It bears minimal overhead while capturing short- and long-term information through a two-level adaptive scheme. To achieve this, the TAM uses a local temporal window and a global aggregation weight.

%TPN, or Temporal Pyramid Network, can be integrated into either 2D or 3D backbones, enhancing model performance, particularly for classes characterized by significant temporal variation. TPN's methodology involves extracting spatial, temporal, and informational features, which are subsequently rescaled and concatenated.

TPN (temporal pyramid network) integrates into 2D or 3D backbones and enhances performance for classes with temporal variation. TPNs extract spatial, temporal, and informational features which are subsequently rescaled and concatenated.

\subsection{Datasets}
\label{sec:datasets} 
        
%We selected three different datasets, Kinetics 400 \cite{kinetics400}, UCF101 \cite{ucf101}, and ETRI-Activity3D \cite{etri}, to assess the explanations on video data. This choice was motivated by the expectation of observing divergent results in both evaluation and explanations, as the datasets exhibit different characteristics.

The Kinetics 400 \cite{kinetics400}, UCF101 \cite{ucf101}, and ETRI-Activity3D \cite{etri} datasets were used to assess the explanations on video data. The goal was to obtain divergent results in both the evaluations and explanations, as the datasets have different characteristics.

%Kinetics 400 is a sizable dataset containing videos sourced from YouTube, encompassing 400 human action classes, with a minimum of 400 clips per class. These videos center on human interactions, encompassing human-human and human-object scenarios, and exhibit a wide range of individuals, environments, and objects. Moreover, the videos often include camera movements and cuts within the same video, introducing complexities in the data.

Kinetics 400 contains 306,245 videos from YouTube containing 400 human action classes with a minimum of 400 clips per class. The videos include human-human and human-object scenarios with a range of individuals, environments, and objects. Moreover, the videos include camera movements and scene cuts, thereby adding complexity to the data.

%UCF101 is a smaller dataset comprising 13,320 videos from 101 classes, all sourced from YouTube. It exhibits significant diversity in terms of camera motion, object appearance and pose, object scale, and illumination. The classes are categorized into five types: human-object interaction, body-motion only, human-human interaction, playing musical instruments, and sports. Additionally, UCF101-24, a subset of this dataset containing only 24 classes, includes action detection annotations from the THUMOS-2013 challenge \citep{THUMOS13}, and was utilized for explanations evaluation purposes.

UCF101 contains 13,320 videos from 101 classes, all sourced from YouTube. The videos are diverse in camera motion, object appearance and pose, object scale, and illumination. There are five class types: human-object interaction, body-only motion, human-human interaction, playing musical instruments, and sports. Additionally, UCF101-24, a subset of this dataset with just 24 classes, includes action detection annotations from the THUMOS-2013 challenge \citep{THUMOS13}; it was used to evaluate explanations.

%On the other hand, ETRI-Activity3D comprises a smaller set of 55 activity classes, with a total of 112,620 videos. These videos capture daily activities of 100 individuals, with half of the participants being above 64 years old, recorded within their home environments from various rooms and eight perspectives. Apart from the RGB videos, depth frames and skeleton joints data are also available. The videos are recorded using fixed cameras and do not include cuts within a single video, offering a more controlled and stable setting for evaluation.

In contrast, ETRI-Activity3D is a smaller set of 55 activity classes from 112,620 videos. The videos capture daily activities of 100 individuals, with half the participants above 64 years old and recorded in their home from various rooms and eight perspectives. Apart from the RGB videos, depth frames and skeleton joints data are also available. The recordings used fixed cameras and do not include scene cuts; they offer a controlled and stable setting for evaluation.

\subsection{Training}

%For training the three networks (TimeSformer, TANet, and TPN), we utilized a computer equipped with a NVIDIA 4090 GPU and an i9 9900KF CPU, kindly provided by the Universitat de les Illes Balears. We conducted the training and evaluation using MMAction2 \cite{2020mmaction2}, an open-source toolbox for video understanding based on PyTorch and part of the OpenMMLab project.

For training the TimeSformer, TANet, and TPN networks we used a computer with an NVIDIA 4090 GPU and an i9 9900KF CPU, kindly provided by the local university. Training and evaluation used MMAction2 \cite{2020mmaction2}, an open-source toolbox for video understanding based on PyTorch and part of the OpenMMLab project.

%Each model was initialized with pre-trained weights from the Kinetics 400 dataset, and further trained for 10 epochs on the ETRI-Activity3D and UCF101 datasets using K-Cross Validation with K=5. The training of the three models on the Kinetics 400 dataset was deemed unnecessary, as pre-trained weights are readily available and publicly accessible online.

The models were initialized with pre-trained weights from the Kinetics 400 dataset, and further trained for 10 epochs on the ETRI-Activity3D and UCF101 datasets using K-cross validation with K~=~5. Training the models on the Kinetics 400 dataset was unnecessary, as pre-trained weights are publicly accessible online.

\subsection{Explanation Methods}
\label{sec:explanation_methods}

%We selected six explanation methods for our experiment: LIME \cite{ribeiro2016why}, Kernel SHAP \cite{lundberg2017unified}, RISE \cite{petsiuk2018rise}, LOCO \cite{lei2018distribution}, Univariate Predictors \cite{guyon2003introduction}, and Occlusion Sensitivity \cite{zeiler2014visualizing}. While LIME, Kernel SHAP, RISE, and Occlusion Sensitivity have already been adapted to image inputs, requiring minimal changes for video input, the remaining two methods, LOCO and Univariate Predictors, needed more substantial modifications to function on video inputs and to provide local explanations instead of global ones. In this section, we detail the adaptations made for each explanation method to work with videos and describe the choices made for each step of the explanation process, summarized in Table \ref{tab:methods}.
        
Our experiment used six explanation methods: LIME \cite{ribeiro2016why}, Kernel SHAP \cite{lundberg2017unified}, RISE \cite{petsiuk2018rise}, LOCO \cite{lei2018distribution}, univariate predictors \cite{guyon2003introduction}, and occlusion sensitivity \cite{zeiler2014visualizing}. While LIME, Kernel SHAP, RISE, and occlusion sensitivity work with image inputs and require minimal change for video input, LOCO and univariate predictors needed substantial modifications for video inputs and provide local rather than global explanations. Table~\ref{tab:methods} details the adaptations for each explanation method to work with videos and itemizes the choices for the steps in feature-removal. Steps and method names are abbreviated for space reasons.

        \begin{table}
            \caption{Step Choices by Video Explanation Method}
            \label{tab:methods}
            \setlength\tabcolsep{3pt}
            \begin{adjustbox}{max width=\textwidth}
            \begin{tabular}{l|llllll}
                \toprule
                
                \textbf{Step} & \textbf{LIME} & \textbf{K.-SHAP} & \textbf{RISE} & \textbf{LOCO} & \textbf{UP} & \textbf{SOS} \\
                \midrule
                
                \textbf{Segment} & SLIC & SLIC & 3D grid & SLIC & SLIC & 3D grid \\
                
                (Regions)& $(\approx200)$ & $(\approx200)$ & $(4\cdot7^2=196)$ & $(\approx200)$ & $(\approx200)$ & $(1183)$ \\
                
                \textbf{Feat. Sel.} & Multiple & Multiple & Multiple & All but one & Only one & All but one \\
                
                \textbf{Sample Sel.} & 1000 & 1000 & 1000 & $\approx200$ & $\approx200$ & $(1183)$ \\

                \textbf{Feat. Rem.} & Blur & Blur & Blur & Blur & Blur & Blur \\
                
                \textbf{Model beh.} & Prediction & Prediction & Prediction & Prediction & Prediction & Prediction \\
                
                \textbf{Summary} & Linear model & Shapley value & Mean\textsuperscript{1} & Difference\textsuperscript{2} & Difference\textsuperscript{2} & Pred. value\textsuperscript{3} \\
                
                \textbf{Visualiz.} & Heat map & Heat map & Heat map & Heat map & Heat map & Heat map \\
                \bottomrule    

                \multicolumn{7}{l}{\vspace{-.75em}}\\
                \multicolumn{7}{l}{\textsuperscript{1} Mean when included} \\
                \multicolumn{7}{l}{\textsuperscript{2} Difference with base case: no perturbations for Video LOCO and all perturbed for Video UP} \\
                \multicolumn{7}{l}{\textsuperscript{3} Prediction when occluded} \\
            \end{tabular}
            \end{adjustbox}
            \footnotetext[1]{Mean when included}
            \footnotetext[2]{Difference with base case: no perturbations for Video LOCO and all perturbed for Video UP}
            \footnotetext[3]{Prediction when occluded}
        \end{table}

%To adapt LIME and Kernel SHAP to videos, the only step requiring modification was segmentation, where pixels of different frames needed to be included in the computation of superpixels. For RISE, we added one dimension to the 2D grid to include temporal information. The LOCO explanations involve multiple trainings of the model, each skipping one feature to perform feature relevance attribution. To avoid repeated retraining, instead we computed the difference between occluding or not each feature (or region) independently, which, when compared to the base prediction (with no occlusions), yielded the relevance of each feature in a local manner. For Univariate Predictors, we changed the method to explain individual predictions (local explanations) rather than the full dataset (global explanations). The relevance of a variable (or region) was found by occluding all but that variable and observing the resulting prediction of the model. For Occlusion Sensitivity, we adapted the 2D occlusion kernel to three dimensions and reduced the number of passes through the model by computing only a sample of all possible occlusions. The occlusion process was regulated by introducing a stride value, determining the number of occlusion possibilities in each dimension, which allowed us to manage computational complexity effectively.

To adapt LIME and Kernel SHAP to videos, the only modification was segmentation, where pixels across frames are used to compute superpixels. For RISE, we added a dimension to the 2D grid for temporal information. The LOCO (leave-one-component-out) explanation method originally requires retraining the model multiple times, excluding a specific feature each time to determine global feature relevance. However, in visual data, global explanations often lack precision due to variability in camera perspective and object positions. To address this, we independently computed the difference between occluding or including each feature (or region), thus yielding the local relevance of each feature. This approach also avoids the need for repeated retraining, a process that is particularly time-intensive with large datasets. The univariate predictors method was adapted to explain individual predictions (local) rather than the full dataset (global). The relevance of a variable (or region) was found by occluding all but that variable and observing the prediction. For occlusion sensitivity, we adapted the 2D occlusion kernel to three dimensions and reduced the passes by using a subset of occlusions. The occlusion process was regulated by a stride value: the number of occlusion possibilities in each dimension. This allowed us to manage computational complexity effectively.

%For methods using superpixels for video segmentation (i.e. LIME, Kernel SHAP, LOCO, and Univariate Predictors), we chose SLIC \cite{achanta2010slic} as the segmentation method due to its ease of extrapolation to three dimensions. We experimentally set the number of segments to approximately 200 and used five times this value as the number of samples for methods requiring this parameter (LIME, Kernel SHAP, and RISE).
        
For methods using superpixels for segmentation (i.e., LIME, Kernel SHAP, LOCO, and univariate predictors), we used SLIC \cite{achanta2010slic} due to its ease of extrapolation to three dimensions. We experimentally set the number of segments to approximately 200 and used $5\times$ this value as the number of samples for methods requiring this parameter (LIME, Kernel SHAP, and RISE).
        
%We experimentally selected the Gaussian blur as the method to apply perturbations to the input. Occlusion with a specific color (e.g. black) was observed to introduce too much noise to the video, and the models could not distinguish the true class, independently of the regions being occluded. Instead, we first computed a blurred version of the input, and introduced its pixels to the perturbed regions. This methodology is more suitable to avoid the problem of generating possible out-of-domain data \cite{miro-nicolau2024assessing}, since perturbations are generated using data already present in the input. Additionally, the masks used to apply the perturbations to the input were also blurred, with a smaller kernel size, to get a ``fade-in'' effect, which allowed the removal of hard edges to the perturbed samples. This ``fade-in'' effect was also used in the visualization of the explanations, to be faithful to the shape of the perturbation each region introduces when occluded.

We experimentally selected a Gaussian blur to apply perturbations to the input, similarly to Fong and Vedaldi \cite{fong2017interpretable}. Occlusion with a specific color (e.g., black) introduced too much noise to the video, preventing the identification of a class independently of the occluded region. Instead, we first computed a blurred version of the input and introduced its pixels to the perturbed regions. This approach is better in avoiding out-of-domain data \cite{miro-nicolau2024assessing}, since perturbations are generated using existing data. Additionally, the masks applied to the input perturbations were also blurred, with a smaller kernel size, to get a ``fade-in'' effect; this allowed hard edges to be removed from the perturbed samples. A fade-in effect was also used in visualizing the explanations, as faithful to the shape of each region when occluded.

%Figure \ref{fig:perturbation-explanation} presents examples of perturbations and explanations for each adapted explanation method. The method names have been slightly modified to reflect their adaption to the video domain and simplified for brevity: Video LIME, Video Kernel-SHAP, Video RISE , Video LOCO, Video UP (UP standing for Univariate Predictors), and Video SOS (SOS standing for Sampled Occlusion Sensitivity to highlight the use of sampling with a stride).
        
\begin{figure}[h]
             \centering
             \includegraphics[width=\textwidth]{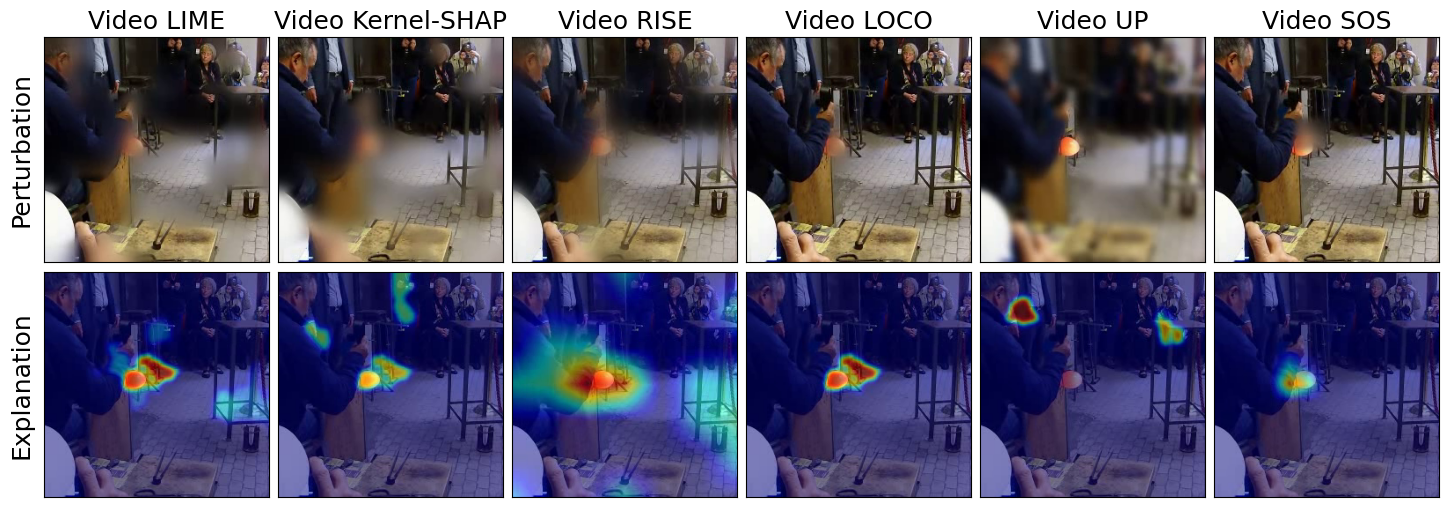}
             \caption{Example explanations by feature removal method. Top row: perturbed sample. Bottom row: saliency map. Explanations were computed for the TANet model on the Kinetics 400 dataset for ``blowing glass.'' While only a single frame is shown, perturbations and explanations extend to the entire video sequence.}
            \label{fig:perturbation-explanation}
        \end{figure}

Figure~\ref{fig:perturbation-explanation} presents perturbations and explanations for each feature removal method. The method names are prefaced with ``Video" to reflect their new context and simplified for brevity: Video LIME, Video Kernel-SHAP, Video RISE , Video LOCO, Video UP (univariate predictors), and Video SOS (sampled occlusion sensitivity, to highlight sampling with a stride).
        
%To enable performance comparison of the proposed methods, we selected two existing model-specific methods, Grad-CAM \cite{selvaraju2017grad} and Saliency Tubes \cite{stergiou2019saliency}, and two model-agnostic ones, AOSA and AOSA-SGL \cite{uchiyama2023visually}. Gradient-weighted Class Activation Mapping (Grad-CAM), is an extended approach that utilizes the gradients from the prediction of a class to weight the features from the last convolutional layer of a Convolutional Neural Network (CNN). Saliency Tubes was inspired by Grad-CAM, but was specifically designed for 3D CNNs, to work on the video domain. Finally, Adaptive Occlusion Sensitivity Analysis (AOSA) was proposed as an extension of \cite{zeiler2014visualizing} to video through the addition of optical flow to extend 2D patches through the temporal dimension.

To compare the methods, we selected two methods that are model-specific (Grad-CAM \cite{selvaraju2017grad} and saliency tubes \cite{stergiou2019saliency}) and two that are model-agnostic (AOSA and AOSA-SGL \cite{uchiyama2023visually}). Gradient-weighted class activation mapping (Grad-CAM) extends to video and uses gradients from the prediction of a class to weight features from the last layer of a convolutional neural network (CNN). Saliency tubes, although inspired by Grad-CAM, was specifically designed for 3D CNNs to work with video. Finally, Adaptive occlusion sensitivity analysis (AOSA) was proposed as an extension to video work by Zeiler and Fergus \cite{zeiler2014visualizing}. It uses optical flow to extend 2D patches through the temporal dimension.

\subsection{Explanation Process}

%To evaluate the proposed explanation methods, we randomly selected 30 videos from each of the three datasets discussed in subsection \ref{sec:datasets}. To ensure consistency, we only included videos that were correctly classified by all three networks used in the experiment.

To evaluate the explanation methods, we randomly selected 30 videos from each dataset discussed in subsection \ref{sec:datasets}. To ensure consistency, we only included videos that were correctly classified by the three networks in the experiment.

        % The selection was not restricted to different classes, allowing for the possibility of having multiple videos from the same class in the sample, while not all classes were necessarily represented. 

%For each dataset, all 30 videos were passed through each of the three networks described in subsection \ref{sec:networks}, and for each prediction, we generated explanations using each of the methods detailed in subsection \ref{sec:explanation_methods} (the six proposed methods, plus four existent ones). This resulted in a total of $30\times3\times3\times10=\text{2,700}$ explanations.

For each dataset, the 30 videos were passed through the three networks described in subsection \ref{sec:networks}, with prediction  explanations generated using each method detailed in subsection \ref{sec:explanation_methods} (the six proposed methods, plus four existing ones). This resulted in  $30\times3\times3\times10 = \text{2,700}$ explanations.

%To ensure consistency across all methods, negative relevance was removed from the explanations, as some methods do not provide both positive and negative relevance values. For visualization, heat maps were employed uniformly across all methods, with important regions represented by warmer colors. Additionally, histogram stretching was applied to maximize the use of the full color range for clearer interpretation.

For consistency, negative relevance was removed from the explanations, as some lack both positive and negative relevance values. For visualization, heat maps were used for all methods, with important regions represented by warmer colors. Additionally, histogram stretching was applied to maximize the color range for clearer interpretation.

\subsection{Evaluation of the Explanations}
\label{sec:evaluation}

%To comprehensively assess the quality of the generated explanations, we mined the literature to find strategies for the evaluation of explanation methods. In this section, we explain in detail the most extended approaches.
        
To assess the quality of the explanations, we mined the literature to find strategies for evaluating explanation methods. In this section, we explain these strategies.
        
\subsubsection{The Deletion and Preservation Games}

%The Deletion and Preservation Games were introduced by Fong et al.\ \citep{fong2017interpretable}, and have been used in three different ways in the literature: to determine minimal masks at target thresholds, to compute the Area Under the Curve (AUC) of varying thresholds, and to estimate the average drop in prediction. These methods were applied to the full set of 2,700 explanations, and are displayed in Figure \ref{fig:insertion-deletion-process}.
    
\begin{figure}[H]
                \captionsetup[subfigure]{justification=centering}
                 \centering
                 \begin{subfigure}[b]{.25\textwidth}
                     \centering
                     \includegraphics[width=\textwidth]{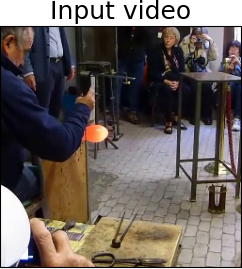}
                     \caption{ }
                    \label{fig:insertion-deletion-input-video}
                 \end{subfigure}
                 \hfill
                 \begin{subfigure}[b]{.25\textwidth}
                     \centering
                     \includegraphics[width=\textwidth]{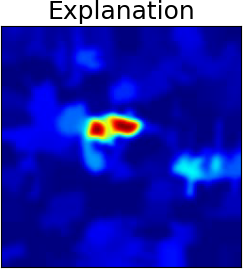}
                     \caption{ }
                    \label{fig:insertion-deletion-explanation}
                 \end{subfigure}
                 \hfill
                 \begin{subfigure}[b]{.25\textwidth}
                     \centering
                     \includegraphics[width=\textwidth]{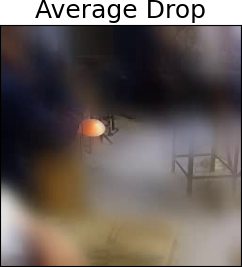}
                     \caption{ }
                    \label{fig:insertion-deletion-avg-drop}
                 \end{subfigure}
                 \begin{subfigure}[b]{1\textwidth}
                     \centering
                     \includegraphics[width=\textwidth]{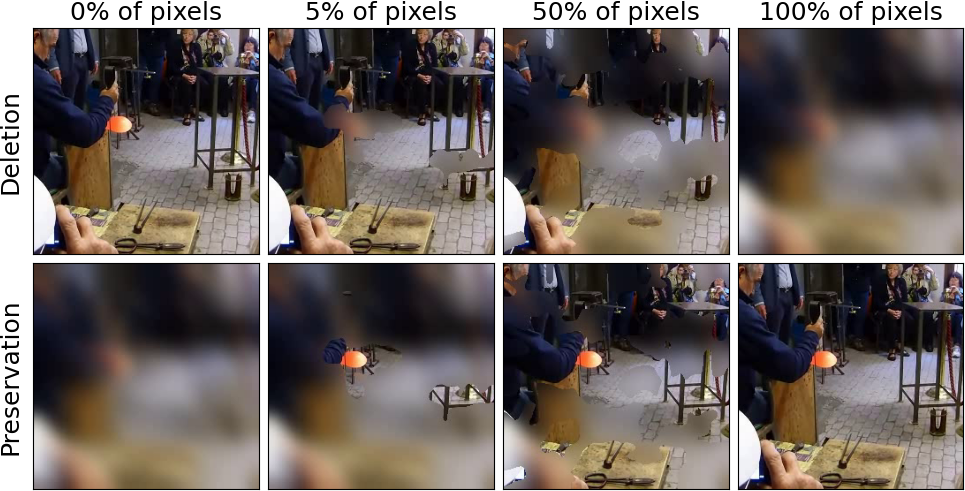}
                     \caption{ }
                    \label{fig:insertion-deletion-examples}
                 \end{subfigure}
                \caption{Perturbations introduced by deleting or preserving regions in the input video, computed on Video LIME explanations for TANet trained on Kinetics 400. Only one frame is shown for simplicity. (a) Input video on ``blowing glass.'' (b) Saliency map, with warmer colors showing greater relevance. (c) Perturbation using the average drop method, with less relevant regions more perturbed. (d) Deletion (top row) and preservation games (bottom row), where regions are removed or preserved according to their relevance. Columns represent perturbations on 0\%, 5\%, 50\%, and 100\% of relevant pixels.}
                \label{fig:insertion-deletion-process}
            \end{figure}

Deletion and preservation games, introduced by Fong and Vedaldi \citep{fong2017interpretable}, are used three ways: to determine minimal masks at target thresholds, to compute the area under the curve (AUC) of varying thresholds, and to estimate the average drop in prediction. Figure \ref{fig:insertion-deletion-process} illustrates our application of these methods to the full set of 2,700 explanations.

%In the context of minimal masks evaluation, the goal is to identify the smallest set of regions that either obstruct sufficient evidence for the network to accurately recognize the correct class (deletion) or provide enough evidence for the correct classification (preservation) \cite{fong2017interpretable}. To accomplish this, minimal masks at various target prediction thresholds are identified, and the number of pixels present in the masks is observed, expressed as a percentage, as models differ in resolution and frame count.
    
For minimal masks evaluation, the goal is to identify the smallest set of regions that either obstruct the network in accurately recognizing the correct class (deletion) or assist the network for the correct classification (preservation) \cite{fong2017interpretable}. For this, minimal masks at target prediction thresholds are identified with the pixels in the masks counted and expressed as a percentage.  Models will differ in resolution and frame count.
    
%Inspired by minimal masks evaluation, Petsiuk et al.\ \citep{petsiuk2018rise} proposed the Area Under the Curve (AUC) of the Deletion and Preservation Games as a metric, which was later used in \citep{petsiuk2021black}. This method involves computing the AUC of the prediction confidence as pixels are deleted or preserved. Unlike the minimal masks evaluation, deletion and preservation AUCs do not require selecting a specific target prediction threshold but rather summarize the results across thresholds. AUC values should be minimal for deletion (with a minimum value of zero) and maximal for preservation (with a maximum value of one). Petsiuk et al.\ referred to the two metrics as ``deletion'' and ``insertion'' metrics, so this naming is adopted for the remaining of the document.

Inspired by minimal masks evaluation, Petsiuk et al.\ \citep{petsiuk2018rise,petsiuk2021black} proposed using the area under the curve (AUC) of the deletion and preservation games as a metric. AUC is applied to the prediction confidence as pixels are deleted or preserved. Unlike the minimal masks evaluation, deletion and preservation AUCs do not use a target prediction threshold but rather summarize the results across thresholds. AUC values range between 0 and 1 and should be minimal for deletion (with a minimum value of 0) and maximal for preservation (with a maximum value of 1). Petsiuk et al.\ referred to the two metrics as ``deletion'' and ``insertion'' metrics, so this naming is adopted for the remaining of the document.

%A variation of the deletion and preservation strategy was proposed in \cite{chattopadhay2018grad}, which they named as the Average Drop, that is, the average decrease in the prediction of a target class when using the original unperturbed video and a perturbed version of it, using the saliency map to determine the regions that should be perturbed. Those regions with more relevance will be lowly perturbed, while regions deemed irrelevant to the model will be perturbed the most. This approach only requires one pass through the network, in contrast to the above methods. Average Drop should be minimal for the explanation method to get a good score.
    
Chattopadhay et al.\ \cite{chattopadhay2018grad} proposed a variation on deletion and preservation called ``average drop'' that measures the average decrease in the prediction when using perturbed and unperturbed video data.  Average drop uses the saliency map to determine the regions to perturb. Regions with more relevance are modestly perturbed, while regions with less relevance are highly perturbed. Only one pass through the network is required. Average drop is minimal for the explanation method to get a good score.
    
%In the three metrics, we utilized the Gaussian blur to remove regions in the deletion metric and as the default value for regions not yet included in the insertion metric. This choice ensures consistency with the perturbations used to compute the explanations, which were generated through the same blurring process.

We utilized the ``deletion'' and ``insertion'' metrics proposed by Petsiuk et al., as they improve upon the original minimal mask evaluation by Fong and Vedaldi \citep{fong2017interpretable} by eliminating the need to select specific thresholds. In addition to these metrics, we included the average drop metric in our experiment. In the three metrics, we utilized a Gaussian blur to remove regions in the deletion metric and as the default value for regions not yet included in the insertion metric. This choice ensures consistency with the perturbations for the explanations, which were generated through the same blurring process.

\subsubsection{Weakly Supervised Object Localization}

%By extracting the most relevant regions that the model deems crucial for the maximum confidence class, it is possible to build a mask for object detection and segmentation without training on a dataset with such labels. This approach has been used to evaluate explanation methods in multiple works \citep{simonyan2013deep, cao2015look, zhang2018top, fong2017interpretable, selvaraju2017grad, chattopadhay2018grad}, but it requires the presence of mask or bounding box ground truth. Consequently, it was computed only on the 900 explanations from the UCF101-24 dataset.

By extracting regions the model deems crucial, it is possible to build a mask for object detection and segmentation without training on a dataset with such labels. This approach is often used  \citep{simonyan2013deep, cao2015look, zhang2018top, fong2017interpretable, selvaraju2017grad, chattopadhay2018grad} but it requires a mask or bounding box for ground truth. Consequently, it was computed only on the 900 explanations from the UCF101-24 dataset.

%In line with the common approach used in prior studies, we employed the Intersection over Union (IoU) metric to evaluate the object localization error for each method, using a threshold of 0.5 \cite{fong2017interpretable, zhang2018top, cao2015look, simonyan2013deep, selvaraju2017grad}. To compute it, we first binarized the saliency maps through simple thresholding, selecting the threshold value that optimized the IoU by sampling across threshold values in the range $[0, 1]$ in 0.05 increments. Additionally, following the methodology in \cite{chattopadhay2018grad}, we performed an evaluation based on the mean IoU value obtained by each method, rather than relying solely on the 0.5 threshold. For the remaining of the document, these two metrics are referred to as ``IoU accuracy'' and ``average IoU''.

In line with prior studies, we used the intersection over union (IoU) metric and a threshold of 0.5 to evaluate the object localization error for each method \cite{fong2017interpretable, zhang2018top, cao2015look, simonyan2013deep, selvaraju2017grad}. To compute it, we binarized the saliency maps through thresholding, and selected the threshold that optimized the IoU by sampling across $[0, 1]$ in 0.05 increments. Additionally, following Chattopadhay et al.'s  methodology \cite{chattopadhay2018grad}, we used the mean IoU for each method, rather than relying solely on the 0.5 threshold. For the remaining of this paper, these two metrics are referred to as ``IoU accuracy'' and ``average IoU''.

%Another commonly used approach for evaluating explanation localization is the Pointing Game \citep{zhang2018top, selvaraju2017grad, petsiuk2018rise, petsiuk2021black}, which assesses whether the maximum point on the saliency map falls within the ground truth bounding box. If the maximum point is located inside the bounding box, it is considered a hit; otherwise, it is classified as a miss. The localization accuracy is then calculated as $Acc = \frac{\text{\#Hits}}{\text{\#Hits}+\text{\#Misses}}$, as defined in \citep{zhang2018top}.

Another approach for evaluating explanation localization is the pointing game \citep{zhang2018top, selvaraju2017grad, petsiuk2018rise, petsiuk2021black}, which determines if the maximum point on the saliency map falls within the ground truth bounding box. If it does, it is considered a hit; otherwise, it is a miss. The localization accuracy is then calculated as $Acc = \frac{\text{\#Hits}}{\text{\#Hits}+\text{\#Misses}}$, as defined by Zhang et al.\ \citep{zhang2018top}.

\begin{figure}[h]
                 \centering
                 \includegraphics[width=1\linewidth]{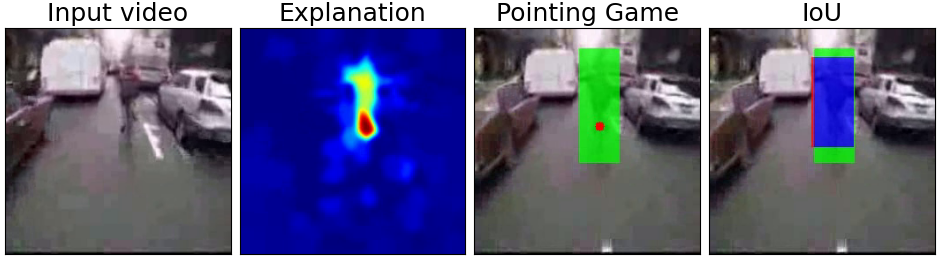}
                 \caption{Comparison of pointing game and intersection of union (IoU).  Left to right: input video, saliency map, pointing game evaluation, and IoU evaluation. Only the frame for maximum relevance is displayed. In the saliency map, relevant regions are in warmer colors. For the pointing game evaluation, the ground truth bounding box is green, while the point of maximum relevance is red. For the IoU evaluation, true positives are blue, false negatives are green, and false positives are red.}
                \label{fig:localization-game}
            \end{figure}

            The two approaches are displayed in Figure \ref{fig:localization-game}.

            % \begin{equation}
            %     Acc = \frac{\text{\#Hits}}{\text{\#Hits}+\text{\#Misses}}.
            %     \label{eq:pointing_game}
            % \end{equation}

\section{Results}

%In this section, we present and analyze the results of the experiment using the evaluation methods outlined in Section \ref{sec:evaluation}. To provide a deeper understanding of the performance of the various methods beyond numerical outcomes, we include multiple examples of explanations across different methods, networks, datasets, and classes, which can be found in Appendix \ref{sec:explanation-examples}.

In this section, we present the results of the experiment using the explanation methods outlined in Section \ref{sec:evaluation}. To provide a deeper understanding beyond numerical outcomes, we include additional examples in Appendix~\ref{sec:explanation-examples} across explanation methods, networks, datasets, and classes.

\subsection{Explanation Time}

%Figure \ref{fig:times} presents the average time required per video for each explanation method. Subfigure \ref{fig:times-model} highlights significant variations in time based on the network architecture. It is important to note that the significance of this metric varies depending on the specific context in which XAI techniques are employed. In the majority of cases, as these techniques are typically applied during the development phase of DL models, rather than in real-time applications, explanation time is generally not a critical concern.
        
\begin{figure}[h]
            \captionsetup[subfigure]{justification=centering}
             \centering
             \begin{subfigure}[b]{.58\textwidth}
                 \centering
                 \includegraphics[width=\textwidth]{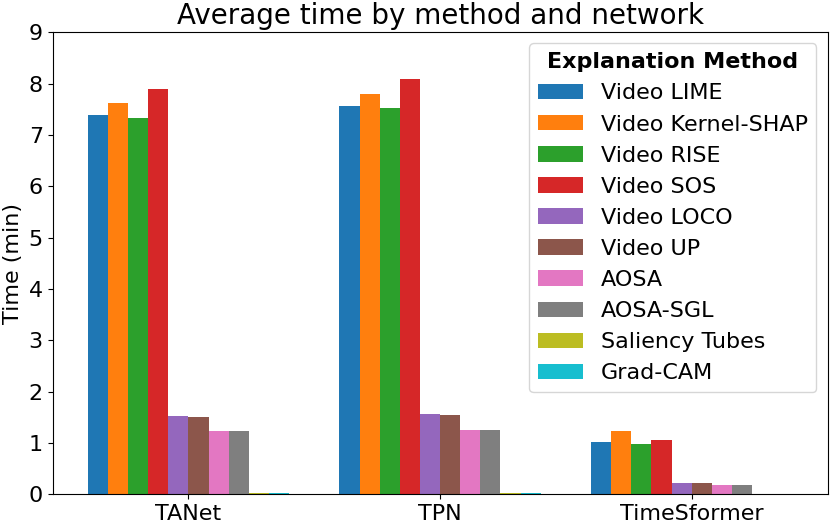}
                 \caption{ }
                \label{fig:times-model}
             \end{subfigure}
             \begin{subfigure}[b]{.41\textwidth}
                 \centering
                 \includegraphics[width=\textwidth]{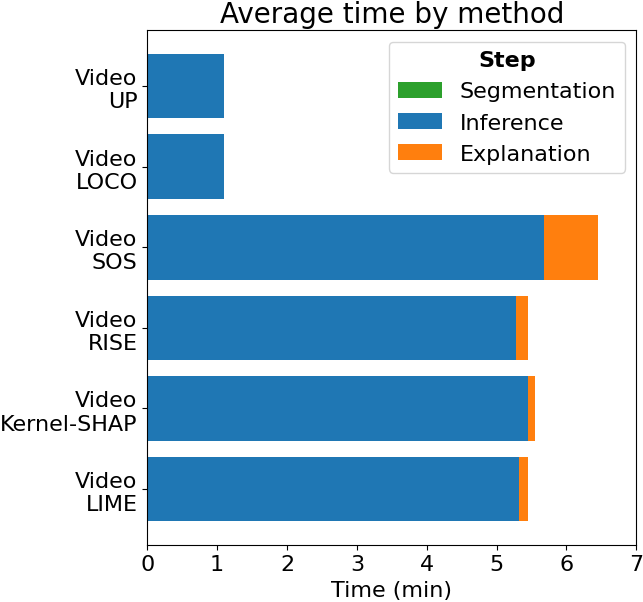}
                 \caption{ }
                \label{fig:times-separated}
             \end{subfigure}
            \caption{Time by method and network.  Input was a single video of resolution 16 $\times$ 112 $\times$ 112 (frames $\times$ width $\times$ height). (a) For comparison, existing methods are included. Bars for saliency tubes and Grad-CAM are barely visible for TANet and TPN as time was under five seconds, and are absent for TimeSformer. (b) Time for the proposed methods by segmentation time (primarily for SLIC computation), inference time (depending on network and sample size), and explanation time (depending on video and sample size).}
            \label{fig:times}
        \end{figure}

This section presents the average time per video for each explanation method. Figure~\ref{fig:times-model} highlights variations in explanation time by network architecture. The significance of time varies by the context in which XAI techniques are employed. Since the techniques are typically applied in developing DL models rather than in real-time use, time is generally not a critical concern.
        
%TANet and TPN, which utilize larger input sizes compared to TimeSformer, require substantially more inference time. This has a considerable impact on methods that necessitate multiple forward passes, such as the proposed approaches. Conversely, methods like Grad-CAM and Saliency Tubes exhibit lower computation times for TANet and TPN, as they require only a single forward and backward pass through the network. However, these methods cannot be applied to TimeSformer due to architectural incompatibility. All methods were evaluated on downscaled versions of the videos (16 $\times$ 112 $\times$ 112) to ensure a fair comparison. This adjustment was necessary due to the significantly higher memory requirements of the official implementations of AOSA and AOSA-SGL\footnote{\url{https://github.com/uchiyama33/AOSA}}, which could not be applied to the full-sized input videos.

TANet and TPN, which use larger input sizes than TimeSformer, require more inference time. This impacts the methods that necessitate multiple forward passes, such as the proposed approaches. Conversely, methods like Grad-CAM and saliency tubes exhibit lower computation times for TANet and TPN, as they require only a single forward and backward pass through the network. However, these methods cannot work with TimeSformer due to architectural incompatibility. All methods were evaluated on down-scaled versions of the videos (16 $\times$ 112 $\times$ 112) to ensure a fair comparison. This was necessary due to the higher memory requirements of AOSA and AOSA-SGL\footnote{\url{https://github.com/uchiyama33/AOSA}}, which preclude using full-size input videos.

%Subplot \ref{fig:times-separated} provides a detailed breakdown of the average time per video, considering all datasets and networks for the proposed methods, and splits the time into different phases: segmentation, inference, and explanation. Although for the use resolution, segmentation time approached zero, for higher resolutions it may be particularly notable for methods that utilize SLIC to partition the video into regions (all proposed methods except Video SOS and Video RISE, which rely on grid-based segmentation). Inference time varies with the network and scales linearly with the number of samples, which differs among explanation methods. Therefore, Video-UP and Video-LOCO, requiring only roughly 200 samples, needed far lesser passes through the model when compared to the remaining methods, requiring roughly 1,000. Finally, the ``explanation time'', encompasses various steps such as feature removal, relevance summary, and visualization. Among these, feature removal is the most computationally expensive step, and its complexity is influenced by factors such as the operations involved, the video size, the number of samples, and the number of regions occluded per sample. Video UP and Video LOCO demonstrated faster feature removal, as they occlude only one region at a time. In contrast, Video Kernel-SHAP, Video LIME, and Video RISE are more time-consuming as they occlude multiple regions simultaneously. Notably, Video-SOS demonstrated the poorest performance, in terms of computation time.

Figure~\ref{fig:times-separated} provides a breakdown of the time per video, considering all datasets and networks for the proposed explanation methods.  Time is split by step; that is, segmentation, inference, and explanation. Although near zero for the given resolution, segmentation time is appreciable for methods that utilize SLIC to partition the video into regions (all proposed methods except Video SOS and Video RISE, which rely on grid-based segmentation). Inference time varies with the network and scales linearly with the number of samples, which differs among explanation methods. Video-UP and Video-LOCO, which require only about 200 samples, needed far fewer passes compared to the remaining methods which required roughly 1,000 passes. Finally, ``explanation time,'' encompasses various steps such as feature removal, relevance summary, and visualization. Among these, feature removal is the most computationally expensive, with complexity influenced by the operations involved, the video size, the number of samples, and the number of regions occluded per sample. Video UP and Video LOCO demonstrated faster feature removal, as they occlude only one region at a time. In contrast, Video Kernel-SHAP, Video LIME, and Video RISE are more time-consuming as they occlude multiple regions simultaneously. Notably, Video-SOS demonstrated the poorest performance.

\subsection{Deletion and Preservation Games}

%Deletion and insertion metrics, calculated using the AUC of the Deletion and Preservation Games using incremental thresholds, are presented in Table \ref{tab:auc}, showing results across various methods, networks, and datasets. Due to certain methods, such as Saliency Tubes and Grad-CAM, being incompatible with TimeSformer's architecture, averaging these values across networks was not possible. In deletion, lower values are preferable, as they indicate a greater drop in model performance when relevant regions are removed compared to irrelevant ones. Conversely, for insertion, higher values are desired, as they reflect a greater improvement in model performance when relevant regions are added back.

Results by explanation method using the AUC of the deletion and preservation games and incremental thresholds are presented in Table~\ref{tab:auc}. The best performance for each network appears in bold. Averaging was not possible for saliency tubes and Grad-CAM, as they are incompatible with TimeSformer's architecture. For deletion, lower values are better as they imply a greater drop in performance when relevant regions are removed compared to irrelevant ones. Conversely, for insertion, higher values are desired as they reflect a greater improvement in performance when relevant regions are added back.

        \begin{table}[h]
        \caption{Deletion and Insertion Scores by Explanation Method and Network}
        \label{tab:auc}
        \begin{adjustbox}{max width=\textwidth}
        \begin{tabular}{l|ccc|ccc}
            \toprule
            \multicolumn{1}{c|}{\multirow{2}{*}{\textbf{Explanation Method}}} & \multicolumn{3}{c|}{\textbf{Deletion (↓)}} & \multicolumn{3}{c}{\textbf{Insertion (↑)}} \\
            \multicolumn{1}{c|}{} & \textbf{TANet}  & \textbf{TPN}    & \textbf{TimeSf.} & \textbf{TANet} & \textbf{TPN} & \textbf{TimeSf.} \\
            \midrule
            \textbf{Video LIME}        & \textbf{0.1694} & \textbf{0.1496} & \textbf{0.1949}      & \textbf{0.7654} & \textbf{0.7646} & \textbf{0.8120}      \\
            \textbf{Video Kernel-SHAP} & 0.2166          & 0.1842          & 0.2741               & 0.7179          & 0.6912          & 0.7400               \\
            \textbf{Video RISE}        & 0.2171          & 0.1788          & 0.2278               & 0.7347          & 0.6703          & 0.7542               \\
            \textbf{Video SOS}         & 0.3800          & 0.2580          & 0.3793               & 0.5923          & 0.5712          & 0.6390               \\
            \textbf{Video LOCO}        & 0.3123          & 0.2421          & 0.3284               & 0.6353          & 0.6568          & 0.7175               \\
            \textbf{Video UP}          & 0.4294          & 0.4042          & 0.5017               & 0.5324          & 0.4607          & 0.5917               \\
            \textbf{AOSA}              & 0.3675          & 0.3503          & 0.4888               & 0.6192          & 0.5756          & 0.6582               \\
            \textbf{AOSA-SGL}          & 0.3786          & 0.3632          & 0.4992               & 0.5911          & 0.5709          & 0.6281               \\
            \textbf{Saliency tubes}    & 0.2190          & 0.1698          & -                  & 0.7167          & 0.7077          & -                  \\
            \textbf{Grad-CAM}          & 0.3177          & 0.1971          & -                  & 0.6416          & 0.6885          & - \\
            \bottomrule
        \end{tabular}
        \end{adjustbox}
        \end{table}

%The results for both metrics show similar trends. The Video LIME method performed best overall, closely followed by Video RISE, Video Kernel-SHAP, and Saliency Tubes, though rankings varied depending on the model and dataset. The poorest results were observed with Video UP, AOSA, AOSA-SGL, and Video SOS. More insights into these two metrics can be found in Figure \ref{fig:auc} in the appendix, displaying results across models and datasets. The best performance for the deletion and insertion metrics were obtained for explanations on the ETRI-Activity3D dataset, with minimum deletion values around 0.1 and insertion values near 0.95. Interestingly, Video LOCO demonstrated relatively strong performance in the insertion metric specifically for the ETRI-Activity3D dataset, outperforming its results in other scenarios.

The results for deletion and insertion show similar trends. The Video LIME method performed best overall, followed by Video RISE, Video Kernel-SHAP, and saliency tubes, though rankings varied depending on network and dataset. The poorest results were observed with Video UP, AOSA, AOSA-SGL, and Video SOS. More insight into these two metrics is found in Figure~\ref{fig:auc} in the appendix. The best performance for deletion and insertion was obtained for explanations on the ETRI-Activity3D dataset, with minimum deletion values around 0.1 and insertion values near 0.95. Interestingly, Video LOCO demonstrated relatively strong performance for insertion, specifically for the ETRI-Activity3D dataset.

%Similar to the deletion metric, lower Average Drop values are more favorable as they indicate a smaller decline in model performance when removing regions based on their relevance (more important regions are preserved better than less important ones). Table \ref{tab:avg-drop} presents these results averaging models, and Figure \ref{fig:avg-drop} in the appendix displays results across models and datasets, similarly to the last two metrics. Unlike the deletion and insertion metrics, a clear top performer emerges: Video RISE, which consistently achieved values below 7\% across all cases, and even under 1\% for the TimeSformer model. 

Similar to deletion, lower average drop values are favorable as they indicate less decline in performance when removing regions based on relevance (i.e., important regions are preserved more often than less important ones). Table~\ref{tab:avg-drop} presents these results by explanation method and network, with the best performance in bold. Note that saliency tubes and Grad-CAM are not applicable to the TimeSformer network. Figure~\ref{fig:avg-drop} in the appendix displays broader results across explanation methods and datasets.  Although Video LIME fared best for deletion and insertion (Table~\ref{tab:auc}), the top performer for average drop was Video RISE with values below 7\% across all cases, and even under 1\% for the TimeSformer network. 

        \begin{table}[h]
        \caption{Average Drop Scores by Explanation Method and Network}
        \label{tab:avg-drop}
        \centering
        \begin{tabular}{l|ccc}
            \toprule
            \multicolumn{1}{c|}{\multirow{2}{*}{\textbf{Explanation Method}}} & \multicolumn{3}{c}{\textbf{Avgerage   Drop (\%, ↓)}} \\
            \multicolumn{1}{c|}{}                                 & \textbf{TANet} & \textbf{TPN} & \textbf{TimeSf.} \\
            \midrule
            \textbf{Video LIME}                                  & 20.51          & 31.88        & 11.19            \\
            \textbf{Video   Kernel-SHAP}                         & 27.22          & 37.73        & 18.60            \\
            \textbf{Video RISE}                                  & \textbf{3.11}           & \textbf{6.10}         & \textbf{0.83}             \\
            \textbf{Video SOS}                                   & 37.93          & 40.16        & 26.76            \\
            \textbf{Video LOCO}                                  & 73.14          & 76.20        & 61.38            \\
            \textbf{Video UP}                                    & 69.19          & 80.98        & 66.70            \\
            \textbf{AOSA}                                        & 34.15          & 22.13        & 21.99            \\
            \textbf{AOSA-SGL}                                    & 41.22          & 25.02        & 25.20            \\
            \textbf{Saliency tubes}                              & 23.50          & 19.10        & -              \\
            \textbf{Grad-CAM}                                    & 29.71          & 24.05        & -             \\
            \bottomrule
        \end{tabular}
        \end{table}

%Video SOS performed well on the Kinetics 400 dataset but demonstrated a significant drop in performance for the other two datasets. Video LIME achieved particularly strong results on the TimeSformer model and maintained consistent, fair results across the other models. AOSA, AOSA-SGL, Saliency Tubes, and Grad-CAM all performed notably well on the TPN model. In contrast, Video LOCO and Video UP consistently delivered the poorest results, showing a significant gap in performance compared to the other methods.

Video SOS performed well on the Kinetics 400 dataset but demonstrated a significant drop in performance for the other two datasets. Video LIME achieved particularly strong results on the TimeSformer model and maintained consistent and fair results across the other models. AOSA, AOSA-SGL, saliency tubes, and Grad-CAM all performed well on the TPN model. In contrast, Video LOCO and Video UP consistently delivered the poorest results, showing a significant gap in performance compared to the other methods.

\subsection{Weakly Supervised Object Localization}

%The results for the weakly supervised object localization metrics, i.e. Pointing Game accuracy, IoU accuracy, and average IoU, were computed solely on the UCF101 dataset, as it was the only one with available bounding box annotations. These results are presented in Tables \ref{tab:pointing-game} and \ref{tab:iou}. Overall, the Pointing Game provided the highest performance, with accuracies ranging between 20\% and 60\%. In contrast, the average IoU values fell between 10\% and 20\%, and IoU accuracy yielded results below 10\% for almost all cases.

Accuracy results for weakly supervised object localization (i.e., pointing game, IoU accuracy, and average IoU) were computed solely on the UCF101 dataset, as it is the only dataset with bounding box annotations. The results are given in Table~\ref{tab:pointing-game} and Table~\ref{tab:iou} with the best results in bold.  Again, saliency tubes and Grad-CAM are not applicable to the TimeSformer network.  Overall, the pointing game provided the highest performance with accuracy between 20\% and 60\%. In contrast, average IoU accuracy was between 10\% and 20\% and IoU accuracy was below 10\% for almost all cases.

        \begin{table}[h]
        \caption{Pointing Game Accuracy by Explanation Method and Network}
        \label{tab:pointing-game}
        \centering
        \begin{tabular}{l|ccc}
            \toprule
            \multicolumn{1}{c|}{\multirow{2}{*}{\textbf{Explanation Method}}} & \multicolumn{3}{c}{\textbf{Pointing Game Accuracy (\%)}} \\
            \multicolumn{1}{c|}{}                                 & \textbf{TANet} & \textbf{TPN} & \textbf{TimeSf.} \\
            \midrule
            \textbf{Video LIME}                                  & 33.33            & 33.33           & 23.33             \\
            \textbf{Video   Kernel-SHAP}                         & 36.67            & 30.00           & 16.67             \\
            \textbf{Video RISE}                                  & \textbf{40.00}   & 33.33           & 33.33             \\
            \textbf{Video SOS}                                   & \textbf{40.00}   & \textbf{53.33}  & 26.67             \\
            \textbf{Video LOCO}                                  & 26.67            & 36.67           & \textbf{43.33}    \\
            \textbf{Video UP}                                    & 33.33            & 23.33           & 20.00             \\
            \textbf{AOSA}                                        & 26.67            & 23.33           & 30.00             \\
            \textbf{AOSA-SGL}                                    & 23.33            & 30.00           & 33.33             \\
            \textbf{Saliency tubes}                              & 30.00            & 26.67           & -               \\
            \textbf{Grad-CAM}                                    & 13.33            & 10.00           & -               \\
            \bottomrule
        \end{tabular}
        \end{table}

        \begin{table}[h]
        \caption{IoU Accuracy and Average IoU by Explanation Method and Network}
        \label{tab:iou}
        \begin{adjustbox}{max width=\textwidth}
        \begin{tabular}{l|ccc|ccc}
            \toprule
            \multicolumn{1}{c|}{\multirow{2}{*}{\textbf{Explanation Method}}} & \multicolumn{3}{c|}{\textbf{IoU Accuracy (\%)}\footnotemark[1]} & \multicolumn{3}{c}{\textbf{Average IoU (\%)}} \\
            \multicolumn{1}{c|}{} & \textbf{TANet}  & \textbf{TPN}    & \textbf{TimeSf.} & \textbf{TANet} & \textbf{TPN} & \textbf{TimeSf.} \\
            \midrule
            \textbf{Video LIME}                                  & \textbf{10.00}          & \textbf{10.00}        & 6.67             & \textbf{17.10}          & \textbf{17.68}        & 15.67            \\
            \textbf{Video   Kernel-SHAP}                         & 6.67           & 6.67         & 6.67             & 16.22          & 16.48        & 15.72            \\
            \textbf{Video RISE}                                  & 6.67           & 6.67         & 6.67             & 15.30          & 15.87        & 16.23            \\
            \textbf{Video SOS}                                   & 0.00           & 3.33         & 3.33             & 11.78          & 12.03        & 13.00            \\
            \textbf{Video LOCO}                                  & \textbf{10.00}          & \textbf{10.00}        & \textbf{16.67}            & 16.54          & 16.23        & \textbf{18.15}            \\
            \textbf{Video UP}                                    & 6.67           & 6.67         & 3.33             & 15.28          & 16.86        & 15.33            \\
            \textbf{AOSA}                                        & 6.67           & 6.67         & 3.33             & 16.63          & 14.08        & 15.38            \\
            \textbf{AOSA-SGL}                                    & 3.33           & 6.67         & 3.33             & 13.51          & 13.37        & 14.52            \\
            \textbf{Saliency tubes}                              & 6.67           & 6.67         & -                & 17.00          & 16.78        & -                \\
            \textbf{Grad-CAM}                                    & 0.00           & 6.67         & -                & 14.66          & 15.30        & -                \\
            \bottomrule
        \end{tabular}
        \end{adjustbox}
        \footnotetext[1]{Threshold of 0.5}        \end{table}
    
%In the Pointing Game, performance varied considerably across models. For the TANet model, Video RISE and Video SOS achieved the highest accuracies (40.00\%), followed closely by Video Kernel-SHAP, Video LIME, and Video UP. On the TPN network, Video SOS stood out with 53.33\% accuracy, significantly outperforming the next best method, Video LOCO, which achieved 36.67\%. In the TimeSformer model, the highest performance was observed with Video LOCO at 43.33\% accuracy, followed by Video RISE and AOSA-SGL, both at 33.33\%. Interestingly, some methods that underperformed in the deletion and insertion tasks, like Video SOS and Video LOCO, demonstrated strong or even the best performance in the localization task.
        
Pointing game performance varied considerably across models. For the TANet network, Video RISE and Video SOS achieved the highest accuracies (40.00\%), followed by Video Kernel-SHAP, Video LIME, and Video UP. On the TPN network, Video SOS stood out with 53.33\% accuracy, significantly outperforming the next best method, Video LOCO at 36.67\%. In the TimeSformer network, the highest performance was observed with Video LOCO at 43.33\% accuracy, followed by Video RISE and AOSA-SGL both at 33.33\%. Interestingly, some methods that underperformed in the deletion and insertion tasks, like Video SOS and Video LOCO, demonstrated strong or even the best performance in the localization task.
        
%When evaluating the methods using IoU-based metrics, including both IoU accuracy and average IoU, the results were relatively consistent across methods. Notable exceptions were Video LOCO, which exhibited superior performance, particularly on the TimeSformer model, achieving an IoU accuracy 10\% higher than the other methods. Conversely, Video SOS consistently showed the poorest results across both metrics in all tested cases.

When evaluating the methods using IoU-based metrics and including both IoU accuracy and average IoU, the results were consistent across methods. Notable exceptions were Video LOCO, which exhibited superior performance, particularly on the TimeSformer model, achieving an IoU accuracy 10\% higher than the other methods. Conversely, Video SOS consistently showed the poorest results across both metrics for all test cases.

\section{Discussion}

%In this section, we discuss the potential of removal-based methods, assess the strengths and weaknesses of the six proposed methods by analyzing the experimental results, and highlight the limitations of the evaluation metrics commonly used in the literature.

In this section, we discuss removal-based methods, assess the strengths and weaknesses of the six proposed methods by analyzing the experimental results, and highlight limitations of evaluation metrics in the literature.

\subsection{Removal-Based Methods}
    
        % Comentar necessitat d'optimitzar hiperparàmetres de mètodes basats en pertorbacions, tant en imatge com en vídeo.
        % Comentar beneficis observats d'aplicar ``smooth masks'' i d'emprar blurring enlloc d'un color específic.

%In the context of removal-based explanations for visual data, it is essential to consider various aspects during the explanation process. This study explored individual steps and discussed key parameters such as segmentation methods, occlusion types, and visualization techniques. This analysis aids in classifying existing explanation methods and paves the way for potential improvements by combining strengths from different methods for various steps.

For removal-based explanations for visual data, it is essential to consider the steps during the explanation process. This study explored individual steps and discussed the parameters that bear on segmentation, occlusion, and visualization technique. This analysis aids in classifying existing explanation methods and paves the way for improvements by combining strengths from different methods and steps.

%We think that the components of each explanation method, e.g. segmentation, feature removal, summary technique, etc., should be analyzed and evaluated separately to find better versions of each method. For instance, the LIME approach of training a linear model to find the relevance of each feature could be applied to the up-scaled grid perturbations generated by RISE. Because there seems to be a lot of different explanation methods, this approach could help to better understand each method, and extract the best of it and disregard the worst. Following the same research direction, the segmentation step should also be studied in more depth: study the impact of the number of regions, of the clustering technique used to compute super-pixels, or even of the relevance of temporal information. Furthermore, by simply generating segmentations focused on a single dimension can potentially lead to explaining them separately, as demonstrated in \cite{gaya_morey2024explainable}. To ensure a fair comparison of the proposed methods, in the performed experiment we enforced equal conditions for the methods, such as the number of features, samples, occlusion type, and visualization. However, the selection of hyper-parameters was performed empirically, leaving the investigation of optimal values for future work. 
        
In our view, the components of explanation methods such as segmentation, feature removal, and summary technique merit separate analysis and evaluation to improve each method. For instance, the LIME approach of training a linear model to find the relevance of each feature could apply to up-scaled grid perturbations generated by RISE. Because there are many explanation methods, this approach offers a better understanding of methods, with follow-up work leveraging the best of it and disregarding the rest. Following the same idea, segmentation merits deeper study, for example, on the impact of the number of regions, clustering for super-pixels, or the relevance of temporal information. Furthermore, segmenting in a single dimension can potentially lead to explaining the dimension separately \cite{gaya_morey2024explainable}. To fairly compare the proposed methods, our experiment enforced equal conditions for the methods, such as the number of features, samples, occlusion type, and visualization. However, the selection of hyper-parameters was performed heuristically, leaving the investigation of optimal values for future work.
        
%A notable discovery during empirical experimentation was the improvement of the resulting explanations when using blurring to introduce perturbations in the input space, when compared to using a specific color, such as black. We believe that this fact is related to the introduction of out-of-domain data into the input when using a specific color \cite{miro-nicolau2024assessing}, while the blurring approach has the advantage of using the data already present, but making it less distinguishable. Furthermore, we also found important the manner in which perturbations are added: introducing them in a ``fade in'' way allows to remove hard edges that could lead to the creation of artificial new shapes, and hence lead to more faithful explanations.

A notable outcome in the experiment was the improvement in explanations using blurring to introduce perturbations in the input space (cf. using a specific color, such as black). This is akin to introducing out-of-domain data when using a specific color \cite{miro-nicolau2024assessing}, while blurring has the advantage of using data already present. Furthermore, the way perturbations are added is important: Using ``fade in'' leads to more faithful explanations and removes hard edges that could lead to artificial new shapes.

\subsection{Proposed Methods: Strengths and Weaknesses}
        % Comentar problema des temps, però avantatge d'agnostic
        % Comentar resultats observats quan comparam a d'altres mètodes, emprant mètodes d'avaluació automàtics.
        % Comentar que explicacions són més precises.
        % Comentar Video SOS vs. AOSA i AOSA-SGL

%Six distinct video explanation methods were proposed, all adapted from existing approaches. The adaptations of Video LIME, Video Kernel-SHAP, and Video RISE were relatively straightforward, as these methods were initially designed for images, requiring only the addition of a temporal dimension during the segmentation step. Our adaptation of Video SOS, however, offers a more natural and robust solution to handle camera movements and cuts, as compared to existing adaptations that rely on optical flow \cite{uchiyama2023visually}. This enables the identification of relevant features in the temporal dimension. Furthermore, we introduced novel adaptations of Video LOCO and Video UP to generate local explanations, extending their applicability to the visual domain beyond global explanations.

Six distinct video explanation methods were proposed, all adapted from existing approaches. Adapting LIME, Kernel-SHAP, and RISE to video was relatively straightforward, as these methods were initially designed for images, and therefore only required adding a temporal dimension during segmentation. The adaptation of Video SOS offers a natural and robust solution to handle camera movements and cuts, as compared to existing adaptations that rely on optical flow \cite{uchiyama2023visually}. This enables the identification of relevant features in the temporal dimension. Furthermore, we introduced novel adaptations, namely Video LOCO and Video UP, to generate local explanations, extending their applicability to the visual domain beyond global explanations.

%The primary limitation of removal-based approaches is their computational cost, as they require multiple perturbed samples and forward passes through the model. However, their significant advantage is being model-agnostic, meaning they can be applied to any model for explanation purposes. This contrast is evident in Figure \ref{fig:times-model}, where Grad-CAM and Saliency Tubes could not be applied to the TimeSformer model, while removal-based methods could be, though the latter required much more computation time. Conversely, the model-specific methods (Grad-CAM and Saliency Tubes) were almost instantaneous for the architectures they could be used on. While Video LOCO and Video UP, though removal-based, were much faster, this came at the cost of less robust results due to the smaller number of samples used.
        
The primary limitation of removal-based approaches is the computational cost, as they require multiple perturbed samples and forward passes through the model. However, they are model-agnostic, meaning they apply to any model. This contrast is evident in Figure~\ref{fig:times-model}, where Grad-CAM and saliency tubes could not work with the TimeSformer model, while removal-based methods could, though the latter required more computation. Conversely, the model-specific methods (Grad-CAM and saliency tubes) were almost instantaneous for the architectures they could be used on. While Video LOCO and Video UP are removal-based and faster, the cost was less robust results due to the smaller number of samples used.
        
%To evaluate the explanation methods, we employed several metrics from the literature, categorized into two primary groups: the Deletion and Preservation Games and weakly supervised object localization. In the Deletion and Preservation Game, Video LIME and Video Kernel-SHAP performed best in terms of deletion and insertion values, while Video RISE outperformed others in the Average Drop metric. Thus, these three methods effectively identified the key regions influencing model performance when removed or added. In contrast, Video SOS excelled in the weakly supervised object localization task, particularly in the Pointing Game, and Video LOCO achieved the best IoU accuracy. The other proposed methods also demonstrated solid results in these metrics, particularly in the Pointing Game, often surpassing Saliency Tubes and Grad-CAM, indicating better localization of relevant regions, notably around human bounding boxes.
        
In evaluating the explanation methods, we employed existing metrics categorized in two groups: the deletion and preservation games and weakly supervised object localization. In the deletion and preservation games, Video LIME and Video Kernel-SHAP performed best for deletion and insertion values, while Video RISE outperformed others in the average drop metric. Thus, these three methods effectively identified the key regions influencing model performance when removed or added. In contrast, Video SOS excelled in the weakly supervised object localization task, particularly in the pointing game, and Video LOCO achieved the best IoU accuracy. The other proposed methods demonstrated solid results in these metrics, particularly in the pointing game, often surpassing saliency tubes and Grad-CAM, indicating better localization of relevant regions, notably around human bounding boxes.
        
%One significant finding was the low performance of the Video UP method, as reflected in multiple metrics. This could be due to excessive perturbations in the explanations, where most samples had all but one region perturbed. As a result, these samples may fall outside the domain, failing to improve confidence in the correct class, even when crucial regions are not perturbed. Additionally, Video Kernel-SHAP explanations scored lower than Video LIME, likely due to the use of numerous samples with over half of the regions occluded, leading to potentially noisy or misleading explanations. Further investigation is warranted to enhance the performance of these two methods.
        
One significant finding was the low performance of the Video UP method, as reflected in multiple metrics. This is due to excessive perturbations in the explanations, where most samples had all but one region perturbed. As a result, these samples may fall outside the domain, failing to improve confidence in the correct class, even when crucial regions are not perturbed. Additionally, Video Kernel-SHAP explanations scored lower than Video LIME, likely due to the number of samples with over half the regions occluded, leading to potentially noisy or misleading explanations. Further investigation is warranted to enhance the performance of these two methods.
        
%When comparing our proposed Video Sampled Occlusion Sensitivity (Video SOS) method to the two versions of Adaptive Occlusion Sensitivity Analysis (AOSA and AOSA-SGL) \cite{uchiyama2023visually}, we observed that Video SOS outperformed in the deletion metric (0.34 vs. 0.40 and 0.41, averaging results across datasets and models) and the Pointing Game (40.00\% vs. 26.67\% and 28.89\%). However, it yielded poorer results in the Average Drop (34.95\% vs. 26.09\% and 30.48\%) and comparable results in the insertion metric (0.60 vs. 0.62 and 0.60), IoU accuracy (0.74\% vs. 1.85\% and 1.48\%), and average IoU (12.27\% vs. 15.36\% and 13.80\%). Notice that for the deletion and Average Drop metrics, lower values are preferred. These results demonstrate the effectiveness of our approach, showing that simply adding the temporal dimension to the occlusion patch can yield accurate explanations without relying on optical flow, which struggles with issues like camera cuts and rapid movements, and completely disregards temporal relevance in the explanations. To further improve Video SOS, both in terms of explanation speed and performance, strategies to include perturbation of multiple regions concurrently should be studied in future work, following the example of many other methods, including AOSA \cite{uchiyama2023visually}.
        
When comparing our proposed video sampled occlusion sensitivity (Video SOS) method to the two versions of adaptive occlusion sensitivity analysis (AOSA and AOSA-SGL) \cite{uchiyama2023visually}, we observed that Video SOS was better in the deletion metric (0.34 vs. 0.40 and 0.41, averaging results across datasets and models) and the pointing game (40.00\% vs. 26.67\% and 28.89\%). However, it yielded poorer results in the average drop (34.95\% vs. 26.09\% and 30.48\%) and comparable results in the insertion metric (0.60 vs. 0.62 and 0.60), IoU accuracy (0.74\% vs. 1.85\% and 1.48\%), and average IoU (12.27\% vs. 15.36\% and 13.80\%). Note that for the deletion and average drop metrics, lower values are preferred. These results demonstrate the effectiveness of our approach, showing that simply adding the temporal dimension to the occlusion patch yields accurate explanations.  This is in contrast to optical flow, which struggles with camera cuts and rapid movements, and completely disregards temporal relevance. To further improve Video SOS for explanation speed and performance, strategies to include perturbation of multiple regions concurrently are topics for future work, following the example of other methods, including AOSA \cite{uchiyama2023visually}.
        
%Lastly, one of the strengths of methods relying on SLIC-based segmentation, such as Video LIME, Video Kernel-SHAP, Video LOCO, and Video UP, is their ability to generate more fine-grained explanations compared to methods like Saliency Tubes and Grad-CAM, which are constrained by the resolution of the final convolutional layer of the network. Additionally, they also surpass methods like Video RISE, Video SOS, AOSA, and AOSA-SGL, which rely on grid-based perturbations. In contrast, superpixel-based segmentation provides more intuitive perturbations, as it tends to include entire objects or uniformly colored regions in the same segment, leading to more interpretable explanations.

Lastly, a strength of methods relying on SLIC-based segmentation, such as Video LIME, Video Kernel-SHAP, Video LOCO, and Video UP, is their ability to generate fine-grained explanations compared to methods like saliency tubes and Grad-CAM, which are constrained by the resolution of the final convolutional layer of the network. Additionally, they also surpass methods like Video RISE, Video SOS, AOSA, and AOSA-SGL, which rely on grid-based perturbations. In contrast, superpixel-based segmentation provides more intuitive perturbations, as it tends to include entire objects or uniformly colored regions in the same segment, leading to more interpretable explanations.

\subsection{Main Problems Found With Existing Evaluation Methods of Explanation Methods}
        % Comentar necessitat de millors mètodes d'avaluació d'explicacions
        % Comentar problemes de localization: no té perquè coincidir amb sa persona: cas de Cricquet bowling. Pointing Game millor que IoU, perquè no és millor cobrir tota sa superfície de sa persona: explicacions menys precises. Video SOS, per exemple, aconsegueix es millor Pointing Game, però es pitjor IoU.
        % Comentar que mètodes amb millors resultats a insertion-deletion pareixen bastant contraris a localization.
        % Comentar problemes de Avg. drop: millors resultats poden ser deguts a que en total hi hagi major importància repartida, i pitjors perquè hi hagi zones molt molt importants comparades amb la resta.
        % Comentar que s'hauria de fer estudi amb usuaris

%To enable a rigorous comparison of the proposed explanation techniques, we adopted multiple evaluation methods widely used in the literature. However, we discovered inherent limitations for some of these methods that should be considered during application. First and foremost, it is important to recognize that the Deletion and Preservation Games and weakly supervised object localization metrics assess entirely different aspects of model explanations. As shown in previous sections, the methods that perform best on these two metrics often differ.

To enable a rigorous comparison of the explanation techniques, we adopted evaluation methods widely used in the literature. However, there are inherent limitations to consider during application. First and foremost, it is important to recognize that the deletion and preservation games and weakly supervised object localization metrics assess entirely different aspects of model explanations. As shown in previous sections, the methods that perform best on these two metrics often differ.

%The Deletion and Preservation Games evaluate how effectively the explanation methods identify the most relevant regions in the input. However, this evaluation relies on perturbations in the input space, which could inadvertently affect model predictions regardless of the region being perturbed. Therefore, we suggest that models' robustness to increasing levels of random perturbation should be computed and factored into the deletion and insertion scores. Additionally, we believe that the Average Drop metric may favor methods with more sparse saliency maps, as these maps result in fewer perturbed regions, thereby leading to smaller decreases in model confidence. This could explain the superior performance of Video RISE on this metric in the experiment compared to other methods, while it was not the case in other metrics.
        
The deletion and preservation games evaluate how effectively the explanation methods identify the most relevant input regions. However, this evaluation relies on perturbations in the input space, which inadvertently affect model predictions regardless of the region perturbed. Therefore, we suggest computing models' robustness to increasing levels of random perturbation and factor that into the deletion and insertion scores. Additionally, we believe the average drop metric may favor methods with more sparse saliency maps, as these result in fewer perturbed regions, thereby leading to smaller decreases in model confidence. This could explain the superior performance of Video RISE on this metric compared to other methods, while it was not the case in other metrics.
        
%In contrast, weakly supervised object localization aims to evaluate the spatial alignment of the most relevant regions in the saliency map, which is primarily influenced by the network itself rather than the explanation method. For this reason, we argue that these metrics are more suitable for comparing trained networks rather than explanation methods. A clear example of this is illustrated in Figure \ref{fig:examples-ucf2}, which depicts the ``cricket bowling'' activity. In this case, the most relevant region identified is the wicket, which is outside the human bounding box. Although this explanation appears reasonable and is consistent across multiple methods, localization metrics would still yield low scores for these explanations. Moreover, we believe that IoU-based metrics overlook focused explanations while favoring more diffuse or widespread explanations. For instance, focusing solely on a person's face may be sufficient for correct classification, but the face's relatively small area compared to the entire bounding box (which encompasses the full body) would result in a low IoU score. Conversely, the Pointing Game would recognize this as a valid explanation. Overall, we believe the Pointing Game is an appropriate metric for evaluating whether models attend to the desired regions (e.g., it could be useful in identifying training flaws in healthcare-related applications). In contrast, IoU-based metrics seem more appropriate for determining whether a trained model is suitable for weakly supervised object localization tasks.
    
In contrast, weakly supervised object localization evaluates the spatial alignment of relevant regions in the saliency map, which is primarily influenced by the network itself rather than the explanation method. For this reason, we argue that these metrics are more suitable for comparing trained networks than explanation methods. A clear example is Figure \ref{fig:examples-ucf2}, which depicts the ``cricket bowling'' activity. Here, the most relevant region is the wicket, which is outside the human bounding box. Although this explanation appears reasonable and is consistent across multiple methods, localization metrics would still yield low scores for these explanations. Moreover, we believe that IoU-based metrics overlook focused explanations while favoring more diffuse or widespread explanations. For instance, focusing solely on a person's face may be sufficient for correct classification, but the face's relatively small area compared to the bounding box (which encompasses the full body) yields a low IoU score. Conversely, the pointing game recognizes this as a valid explanation. Overall, we believe the pointing game is an appropriate metric for determining if models attend to the desired regions (e.g., it could be useful in identifying training flaws in healthcare applications). In contrast, IoU-based metrics are more appropriate for determining whether a trained model is suitable for weakly supervised object localization tasks.
    
\section{Conclusions}

%As the application of AI expands across various domains, the importance of explainable AI techniques has become increasingly evident, but while numerous approaches have been developed for image data, the video domain remains insufficiently explored. Therefore, in this study, we proposed to decompose the explanation process of removal-based methods into well-defined steps, specifically aimed at handling visual data. This steps are: segmentation, feature selection, sample selection, feature removal, model behavior, summary technique, and visualization. This distinction in the pipeline is crucial when dealing with video data due to the increased complexity resulting from the addition of the temporal dimension. Leveraging this framework, we adapted six removal-based explanation methods to the video domain, incorporating necessary adjustments in each case.
    
As the applications of AI expand, the importance of explainable AI (XAI) techniques is increasingly evident, but while numerous approaches exist for image data, the video domain is insufficiently explored. Therefore, in this research, we decomposed the explanation process of removal-based methods into well-defined steps, specifically aimed at visual data. The steps are segmentation, feature selection, sample selection, feature removal, model behavior, summary technique, and visualization. This distinction in the pipeline is crucial when dealing with video data due to the added temporal dimension. Leveraging this framework, we adapted six removal-based explanation methods to the video domain, incorporating necessary adjustments in each case.
    
%The evaluation encompassed metrics related to the Deletion and Insertion Game, and to the weakly supervised localization task. The results reveal consistent good results of the proposed methods across the different metrics. More precisely, Video LIME, Video Kernel-SHAP and Video RISE showed better performance for the deletion and insertion metrics than model-specific methods like Grad-CAM and Saliency Tube, and than other video-specific removal-based approaches, like AOSA. Video SOS and Video LOCO, on the other hand, excelled in the localization metrics.
    
The evaluations used metrics for the deletion and insertion game and the weakly supervised localization task. The results were consistently good for the proposed methods across the different metrics. Specifically, Video LIME, Video Kernel-SHAP and Video RISE showed better performance for the deletion and insertion metrics than model-specific methods like Grad-CAM and saliency tubes, and other video-specific removal-based approaches, like AOSA. Video SOS and Video LOCO excelled in the localization metrics.
    
%Decomposing the explanation process into well-defined steps presents opportunities for investigating the influence of each choice along the explanation pipeline, ultimately enhancing the quality of explanation methods. Future research endeavors should explore combining aspects from different methods to derive optimal synergies, and a wider range of choices for each step to optimize the methods and better understand how different kinds of perturbations affect models' predictions.

Decomposing the explanation process into well-defined steps presents opportunities for investigating the influence of each choice in the explanation pipeline, ultimately improving the explanation methods. Future research should combine aspects from different methods to derive optimal synergies and choices for each step to optimize the methods and better understand how different perturbations affect model predictions.

\appendix

\section{Examples of explanations}
\label{sec:explanation-examples}

%Figures \ref{fig:examples-etri1}, \ref{fig:examples-etri2}, \ref{fig:examples-kinetics1}, \ref{fig:examples-kinetics1}, \ref{fig:examples-ucf1}, and \ref{fig:examples-ucf2} present representative examples of model explanations. These figures illustrate the explanations generated by all methods used in this study for predictions made by three different networks: TANet, TPN, and TimeSformer. Two classes from each dataset—ETRI-Activity3D, Kinetics 400, and UCF101—are displayed.

\begin{figure}[H]
         \centering
         \includegraphics[width=1.\textwidth]{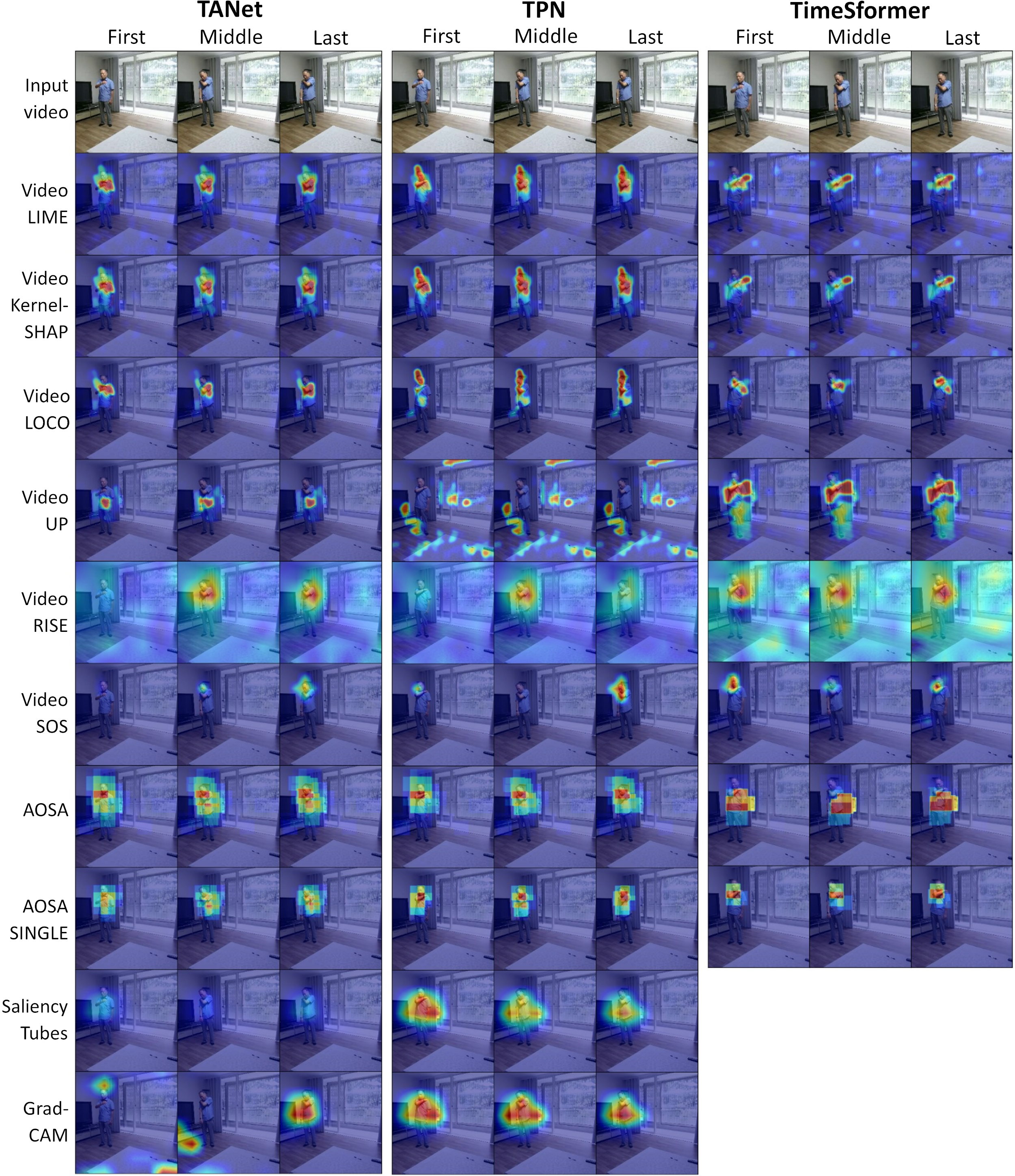}
         \caption{ Explanations for the ETRI-Activity3D class ``massaging a shoulder oneself.'' }
        \label{fig:examples-etri1}
    \end{figure}

\begin{figure}[H]
         \centering
         \includegraphics[width=1.\textwidth]{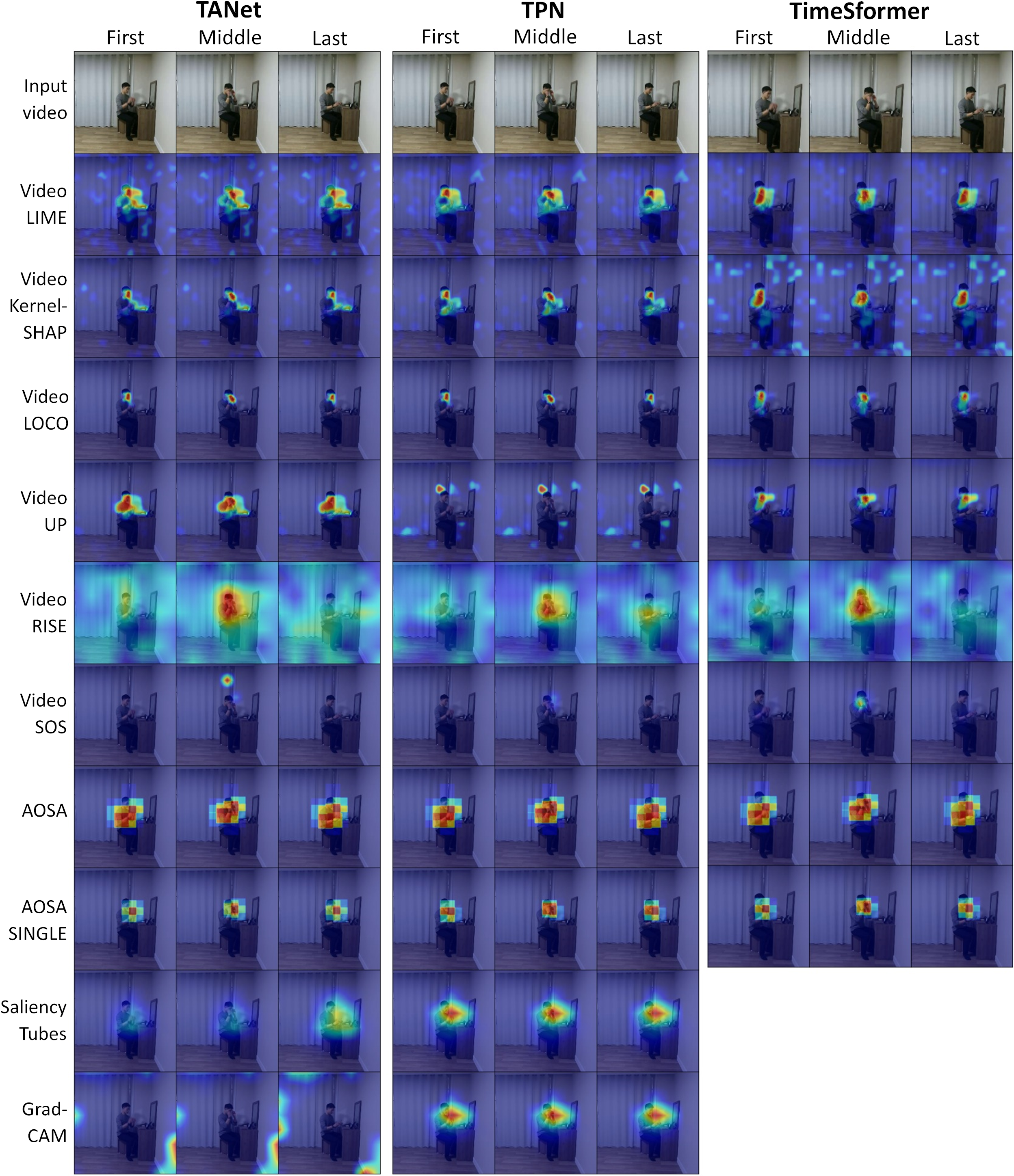}
         \caption{ Explanations for the ETRI-Activity3D class ``putting on/taking off glasses.'' }
        \label{fig:examples-etri2}
    \end{figure}

\begin{figure}[H]
         \centering
         \includegraphics[width=1.\textwidth]{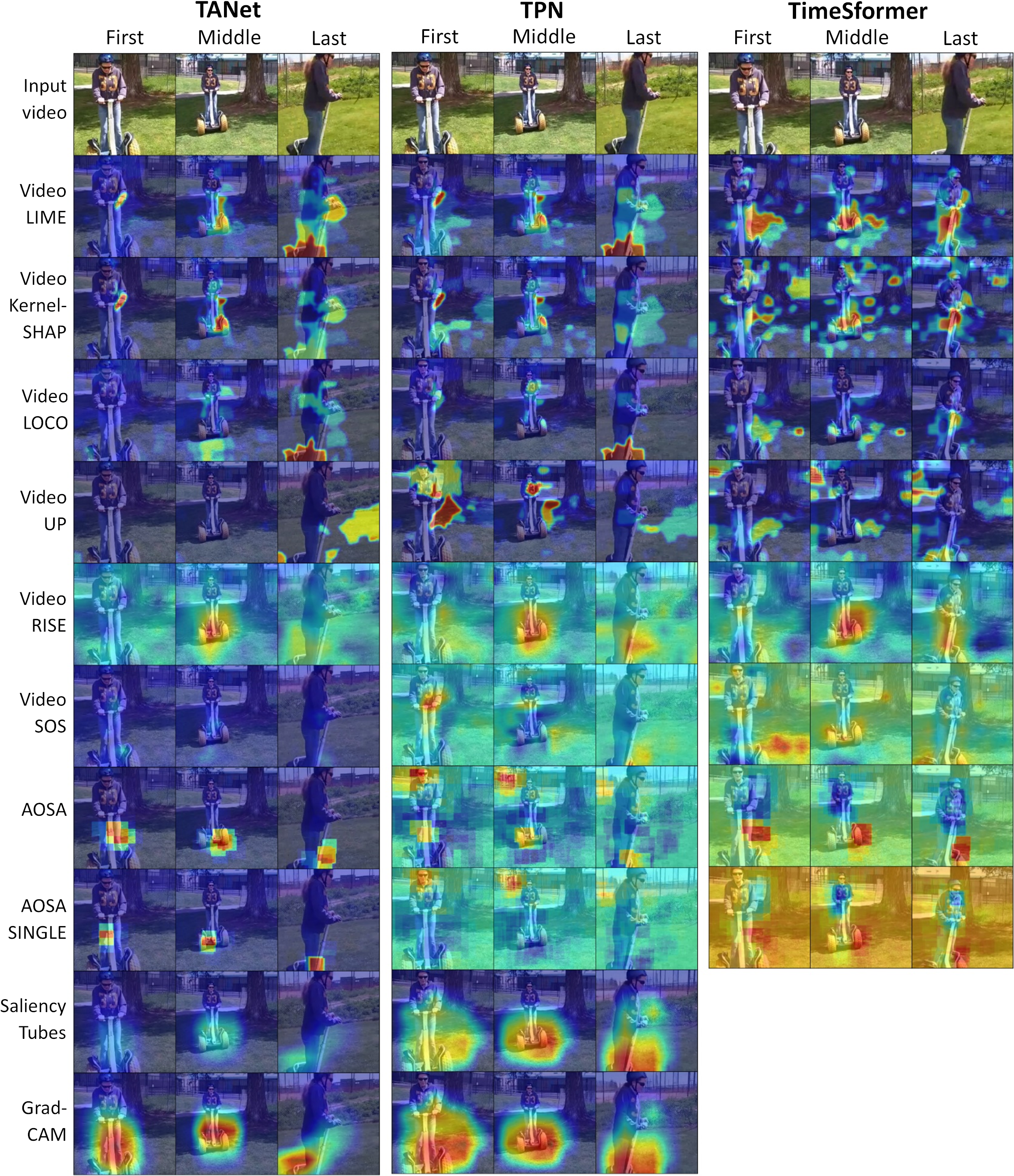}
         \caption{ Explanations for the Kinetics 400 class ``using segway.'' }
        \label{fig:examples-kinetics1}
    \end{figure}

\begin{figure}[H]
         \centering
         \includegraphics[width=1.\textwidth]{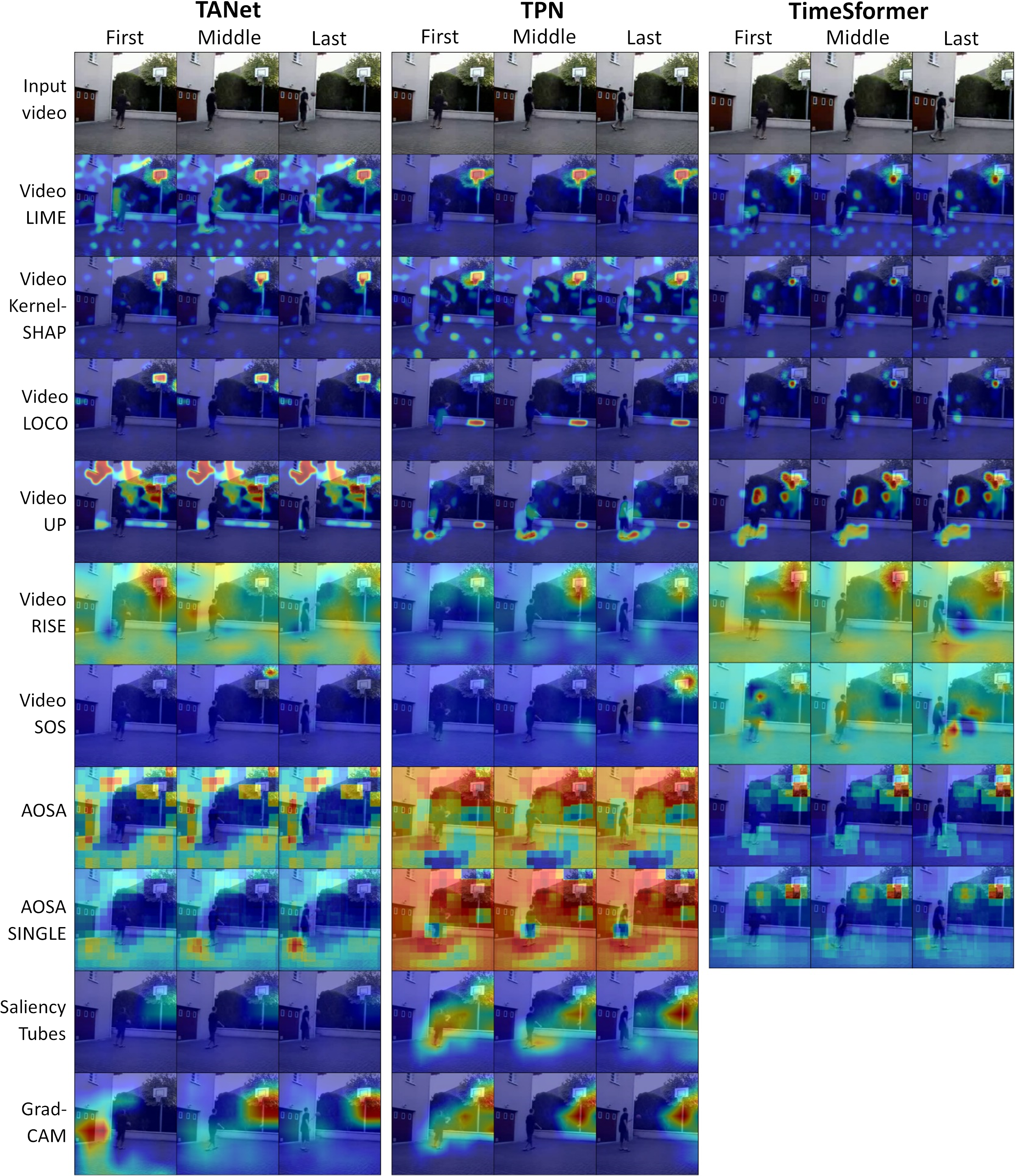}
         \caption{ Explanations for the UCF101 class ``basketball.'' }
        \label{fig:examples-ucf1}
    \end{figure}

\begin{figure}[H]
         \centering
         \includegraphics[width=1.\textwidth]{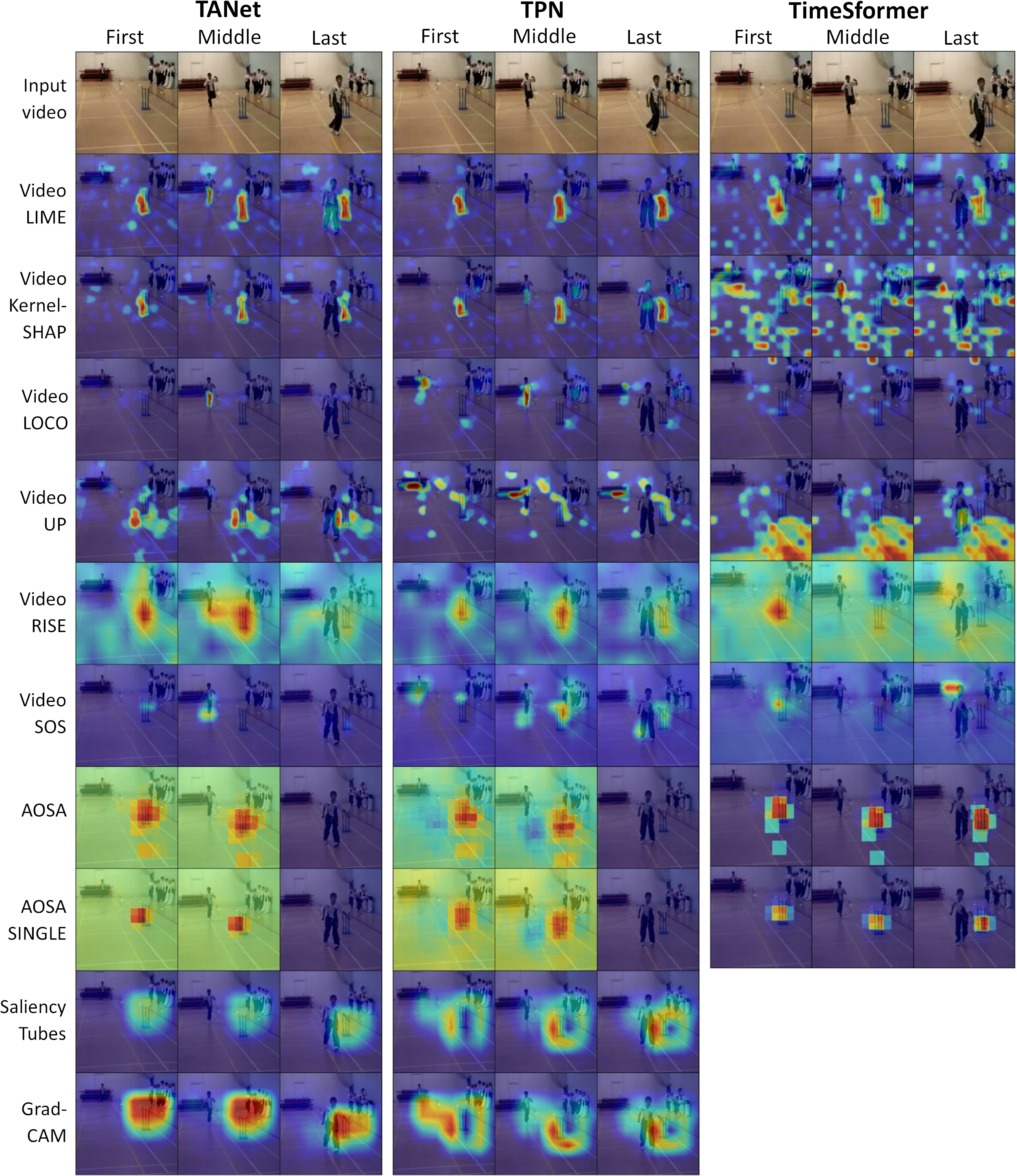}
         \caption{ Explanations for the UCF101 class ``cricket bowling.'' }
        \label{fig:examples-ucf2}
    \end{figure}

Figures \ref{fig:examples-etri1}, \ref{fig:examples-etri2}, \ref{fig:examples-kinetics1}, \ref{fig:examples-kinetics1}, \ref{fig:examples-ucf1}, and \ref{fig:examples-ucf2} present examples of model explanations. The figures illustrate the explanations from all methods in this study for the TANet, TPN, and TimeSformer networks. Two classes from each dataset---ETRI-Activity3D, Kinetics 400, and UCF101---are displayed.

%Each figure is organized as follows: vertically arranged rows represent the input video and the corresponding explanations generated by the six proposed methods, followed by four additional methods used for comparison purposes. Horizontally, the content is divided into three groups, each containing three columns, corresponding to results from the three networks, for three sampled frames, respectively. The frames were sampled at consistent intervals (first, middle and last frames) for visual clarity and comparison.

Each figure is organized as follows: Rows show the input video followed by rows for the six proposed methods and four additional methods for comparison. Columns are divided into three groups (one per network), each containing three columns (first, middle, last) for frames sampled at consistent intervals for comparison.

    \begin{figure}[H]
         \centering
         \includegraphics[width=1.\textwidth]{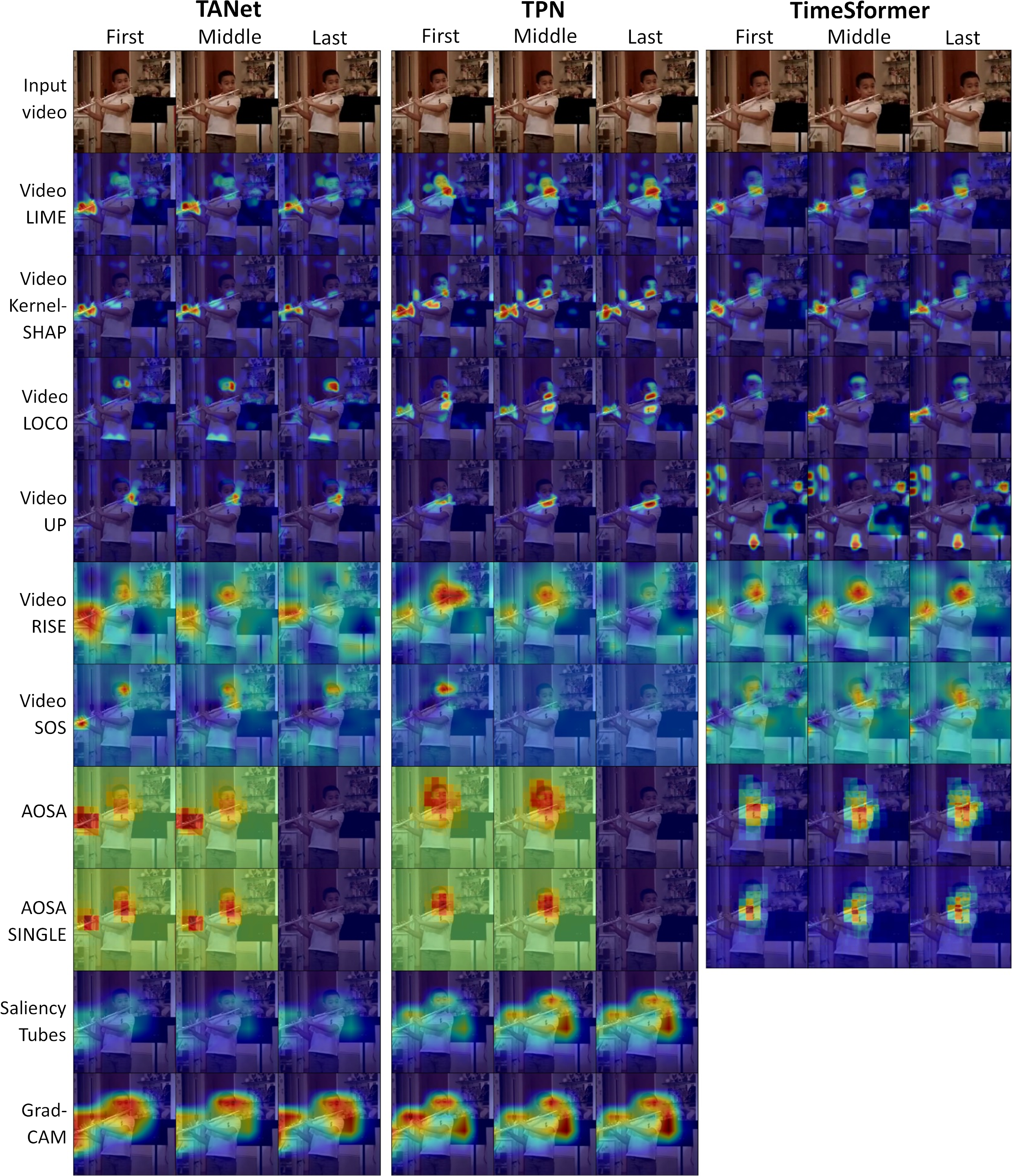}
         \caption{ Explanations for the Kinetics 400 class ``playing flute.'' }
        \label{fig:examples-kinetics2}
    \end{figure}

\section{Results by dataset and model}
\label{sec:results-by-dataset-and-model}

\begin{figure}[H]
         \centering
         \includegraphics[width=\textwidth]{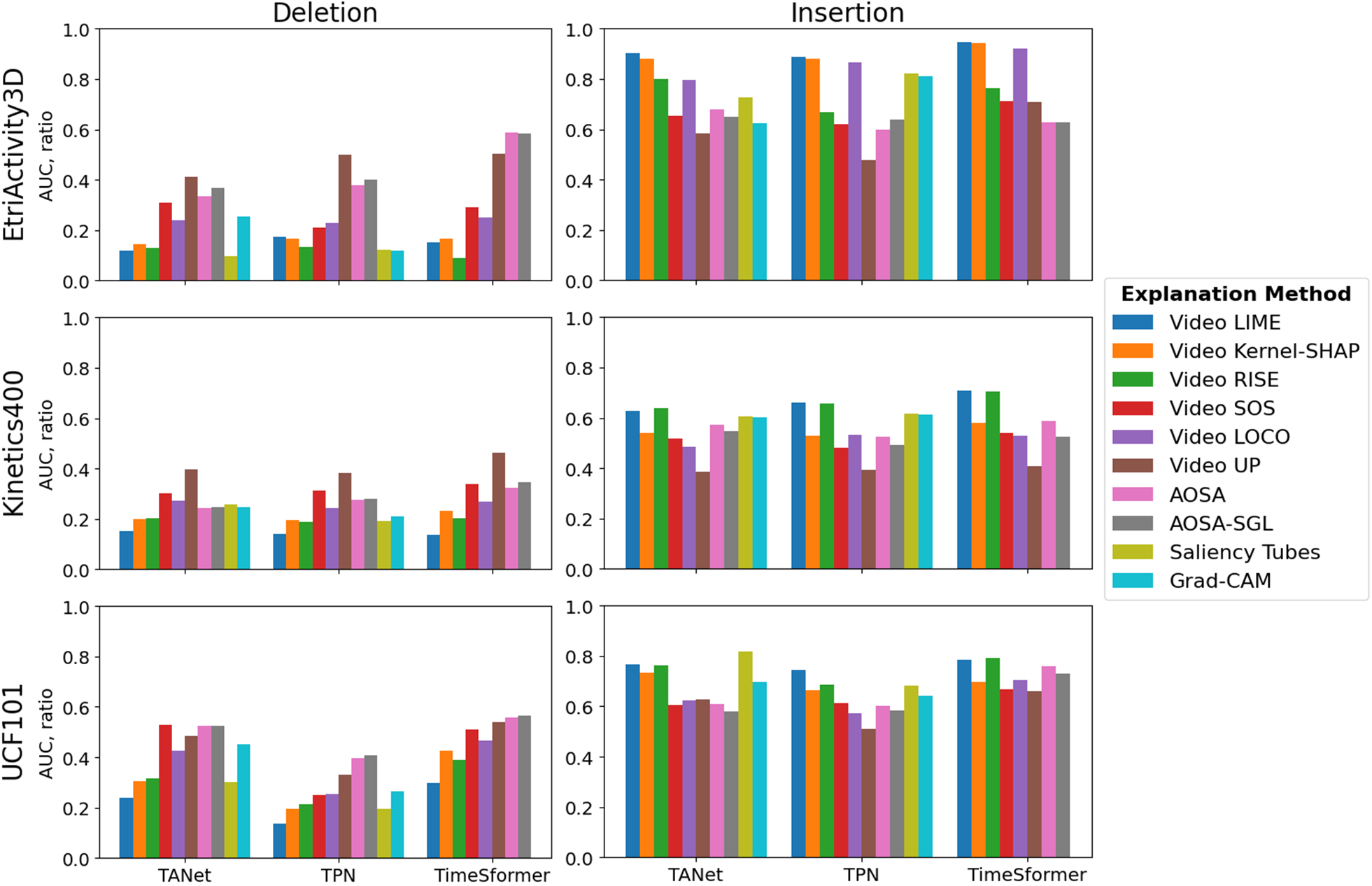}
         \caption{AUC (ratio) by method, network, and dataset. In the left column, deletion game (lower values are better); in the right one, insertion game (higher values are better). By rows, the datasets (ETRI-Activity3D, Kinetics 400, and UCF101) and within each plot, the networks (TANet, TPN, and TimeSformer) and explanation methods (see legend).}
        \label{fig:auc}
    \end{figure}

\begin{figure}[H]
         \centering
         \includegraphics[width=\textwidth]{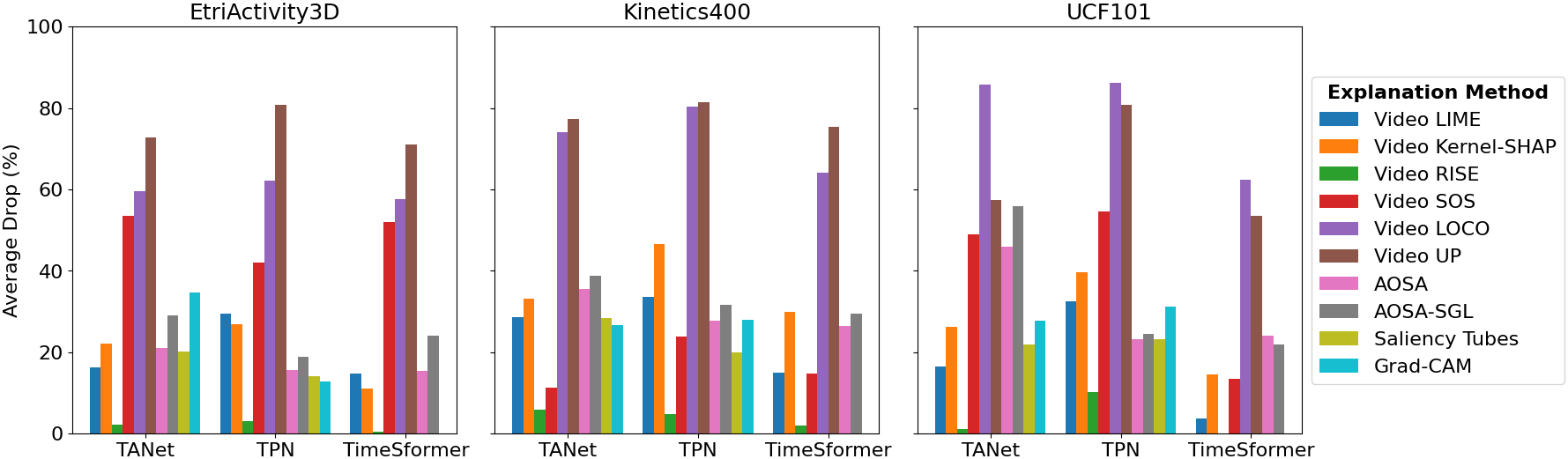}
         \caption{Average drop (\%) by dataset, network, and method (lower is better). By columns, the datasets, ETRI-Activity3D, Kinetics 400, and UCF101, and within each plot, the networks (TimeSformer, TANet and TPN) and explanation methods (see legend).}
        \label{fig:avg-drop}
    \end{figure}

This section provides further insight into the deletion and insertion metrics.  See Figure~\ref{fig:auc} and Figure~\ref{fig:avg-drop}. For both metrics, results are grouped by network, dataset, and method.
        
\section*{CRediT authorship contribution statement}

\textbf{F. Xavier Gaya-Morey:} Conceptualization, Methodology, Software, Validation, Formal Analysis, Investigation, Data Curation, Writing - Original Draft, Writing - Review \& Editing Preparation, Visualization, Project administration. \textbf{Jose M. Buades-Rubio:} Conceptualization, Methodology, Formal Analysis, Writing - Review \& Editing Preparation, Supervision, Project administration, Funding acquisition. \textbf{I. Scott MacKenzie:} Writing - Review \& Editing Preparation, Supervision. \textbf{Cristina Manresa-Yee:} Conceptualization, Methodology, Writing - Review \& Editing Preparation, Supervision, Project administration, Funding acquisition.

\section*{Funding sources}

This work is part of the Project PID2023-149079OB-I00 (EXPLAINME) funded by MICIU/AEI/10.13039/ 501100011033/ and FEDER, EU and of Project PID2022-136779OB-C32 (PLEISAR) funded by MICIU/AEI/10.13039 /501100011033/ and FEDER, EU. F. X. Gaya-Morey was supported by an FPU scholarship from the Ministry of European Funds, University and Culture of the Government of the Balearic Islands.

\section*{Data availability}

The REVEX framework is made publicly available at \url{https://github.com/Xavi3398/revex_framework}.

\section*{Declaration of generative AI and AI-assisted technologies in the writing process}

During the preparation of this work the authors used ChatGPT in order to improve the readability and language of the manuscript. After using this tool, the authors reviewed and edited the content as needed and take full responsibility for the content of the published article.

\bibliographystyle{elsarticle-num}
\bibliography{bibliography}% common bib file


% %% For numbered reference style

%   \textit{\LaTeX: a document preparation system},
%   Addison Wesley, Massachusetts,

% \end{thebibliography}
\end{document}